\documentclass[journal]{IEEEtran}
\usepackage{times}
\usepackage{amsmath,amsfonts}
\usepackage{algorithmic}
\usepackage{algorithm}
\usepackage{array}
\usepackage[caption=false,font=normalsize,labelfont=sf,textfont=sf]{subfig}
\usepackage{textcomp}
\usepackage{stfloats}
\usepackage{url}
\usepackage{verbatim}
\usepackage{graphicx}
\usepackage[nocompress]{cite}
\usepackage{makecell}
\usepackage{multirow}
\usepackage{tabularx}
\usepackage{booktabs} 
\usepackage{lipsum}
\usepackage{amssymb}
\usepackage{xcolor}

\begin{document}

\title{MUSCLE-NET: Predicted-Multiscale-Aware Network for Pedestrian Trajectory Forecasting}

\author{ Yu Liu,~\IEEEmembership{Graduate Student Member,~IEEE,}  Ming Huang, Xiao Ren,~\IEEEmembership{Student Member,~IEEE,} 
Zhijie Liu,~\IEEEmembership{Student Member,~IEEE,}
   Youfu Li,~\IEEEmembership{Fellow,~IEEE,} and He Kong,~\IEEEmembership{Member,~IEEE}

\thanks{This manuscript has been accepted to the IEEE Transactions on Intelligent Transportation Systems as a regular paper. This work was supported by the National Key R$\&$D Program of China under Grant No. 2024YFB4710900, the National Natural Science Foundation of China (NSFC) under Grant No. U24A20265, the Science, Technology, and Innovation Commission of Shenzhen Municipality, China, under Grant No. JCYJ20240813094212017, the Shenzhen Science and Technology Program under Grant No. KQTD20221101093557010, the Guangdong Science and Technology Program under Grant No. 2024B1212010002. (Corresponding author: He Kong)}

\thanks{Yu Liu, Ming Huang, Xiao Ren, Zhijie Liu, and He Kong are with the Guangdong Provincial Key Laboratory of Fully Actuated System Control Theory and Technology, School of Automation and Intelligent Manufacturing, Southern University of Science and Technology (SUSTech), Shenzhen 518055, China. Yu Liu is also with the Department of Mechanical Engineering, City University of Hong Kong, Hong Kong SAR, China. Emails: yuliu254-c@my.cityu.edu.hk; \{12532857, 12431359, 12332642\}@mail.sustech.edu.cn; kongh@sustech.edu.cn. You-Fu Li is with the Department of Mechanical Engineering, City University of Hong Kong, Hong Kong SAR, China. Email: meyfli@cityu.edu.hk.}
}

\markboth{IEEE TRANSACTIONS ON INTELLIGENT TRANSPORTATION SYSTEMS}%
{Shell \MakeLowercase{\textit{et al.}}: A Sample Article Using IEEEtran.cls for IEEE Journals}

\IEEEpubid{\begin{minipage}{\textwidth}\ \centering
    1558-0016\copyright~2024 IEEE. Personal use is permitted, but republication/redistribution requires IEEE permission. \\
    See https://www.ieee.org/publications/rights/index.html for more information.
\end{minipage}}

\maketitle

\begin{abstract}

Accurate pedestrian trajectory prediction is essential for safe navigation in autonomous driving and intelligent transportation systems. Despite substantial progress made by recent methods, most existing approaches are limited in fully exploiting diverse observations and often overlook the scale dependency of future motion, treating multiscale features uniformly regardless of underlying motion dynamics. This limits their robustness across diverse pedestrian behaviors. To address these challenges, we propose a Predicted-MUltiSCale-Aware Network (MUSCLE-NET) for Pedestrian Trajectory Forecasting that integrates complementary multimodal cues with scale-adaptive prediction mechanisms. The proposed framework is built upon a Multiscale Multimodal Feature Extraction (MMFE) module, which combines multiscale representation, modality-aware recalibration, and directional cross-modal fusion to construct semantically aligned representations from bounding boxes, velocities, and pose information. Building on these features, a Multiscale Enhanced Hierarchical Prediction (MEHP) module performs prediction-aware future-motion refinement via a probabilistic coarse predictor, scale-aligned fusion, and progressive refinement, adaptively selecting scale-relevant cues to mitigate spatial drift. Extensive experiments on the JAAD and PIE benchmarks demonstrate that the proposed MUSCLE-Net achieves competitive performance and consistent gains compared with state-of-the-art trajectory prediction methods.

\end{abstract}


\section{Introduction}

\IEEEPARstart{W}ith the rapid advancement of intelligent sensing and computing technologies, significant progress has been made toward the development of autonomous vehicles that enhance traffic efficiency and road safety. A critical requirement for collision avoidance is effective path planning \cite{ref4,ref124}, which relies on understanding interactions among road users and accurately forecasting their future behaviors \cite{ref1}. Given observed traffic scenes and historical motion data, pedestrian trajectory prediction aims to forecast plausible future paths within a short time horizon. Despite recent progress, this task remains challenging due to two key factors \cite{ref27}.
\begin{figure}[t]
\centering
\includegraphics[scale= 0.40]{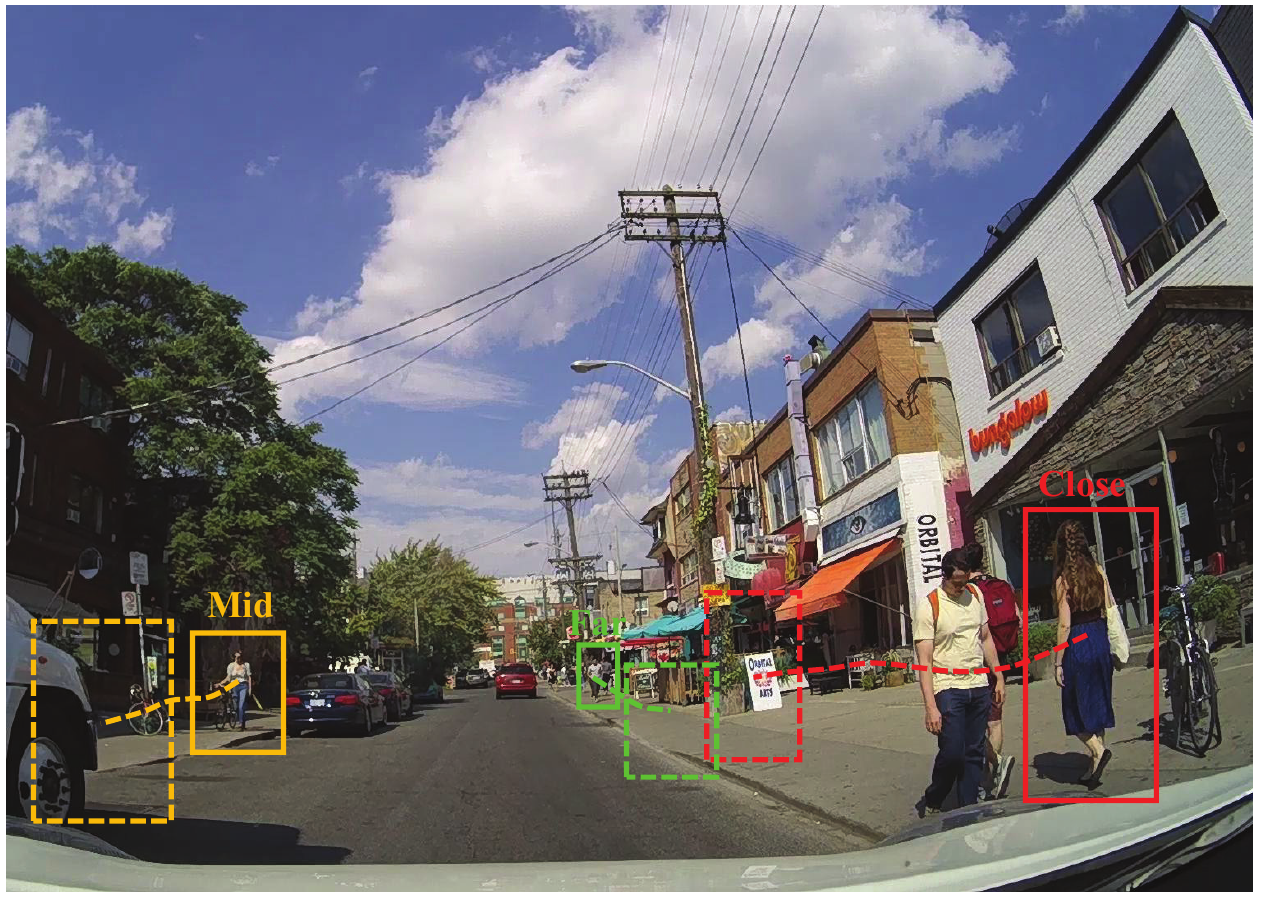}
\caption{Representative examples showing pedestrians at different viewing distances, including Close, Mid, and Far scenarios. The large variation in bounding-box scale, visual clarity, and motion patterns highlights the challenges of scale-aware and robust pedestrian trajectory prediction.}
\centering
\label{intro1}
\end{figure}

\IEEEpubidadjcol  

Pedestrian motion is influenced by diverse and complementary sources of information, including visual appearance, ego vehicle dynamics, bounding box evolution, and body pose cues. Although recent studies \cite{ref104, ref105, ref107, ref108} leverage multimodal inputs to enrich motion representations, different modalities often exhibit varying reliability across time and scenarios, which may introduce noisy, misaligned, or even contradictory features. Moreover, cross-modal relationships are asymmetric. For instance, pose cues often help refine subtle short-term motion, whereas bounding box dynamics and ego-motion signals are more effective for capturing long-term trajectory trends. Most existing prediction methods \cite{ref79, ref66} mainly leverage multimodal cues through feature incorporation and fusion, without explicitly modeling their time-varying reliability, cross-modal inconsistency, or directional dependencies. As a result, constructing reliable and semantically consistent multimodal representations for trajectory prediction remains a challenging problem.

Another key challenge stems from the inherently probabilistic and multimodal nature of pedestrian motion. To model this uncertainty, recent works adopt multiscale representations that capture motion patterns over different temporal extents. However, most existing multiscale approaches \cite{ref96, ref97} treat features from different scales uniformly during prediction, without considering scale dependency in future motion. For example, future motion may evolve toward different scale characteristics during prediction. Pedestrians whose future trajectories move closer often exhibit larger displacement and finer local variations, which benefit from fine-scale features, whereas those whose future trajectories remain distant or move farther away tend to follow smoother motion patterns that are better modeled by coarse-scale representations, as illustrated in Figure~\ref{intro1}. Ignoring such future scale variation limits the effectiveness of multiscale modeling and degrades prediction performance under diverse motion patterns.

Motivated by the above challenges, this work proposes MUSCLE-Net, a Predicted-Multiscale-Aware Network for Pedestrian Trajectory Forecasting, to address multimodal uncertainty and scale-dependent motion prediction within a unified framework. The core design principle of MUSCLE-Net is to integrate calibrated multimodal representation learning and prediction-aware future-motion refinement within a unified framework, rather than treating multimodal and multiscale information uniformly throughout the pipeline. To this end, the architecture comprises two tightly coupled modules. The Multiscale Multimodal Feature Extraction (MMFE) module constructs robust and semantically aligned motion representations from observations through multiscale temporal modeling, adaptive feature recalibration, and directional cross-modal interaction. Building on these representations, the Multiscale Enhanced Hierarchical Prediction (MEHP) module performs prediction-aware multiscale reasoning via probabilistic coarse motion estimation, aligned scale fusion with temporal modulation, and progressive refinement. Together, the two modules enable calibrated multimodal representation learning and prediction-aware multiscale future refinement in a unified framework, leading to more robust and accurate pedestrian trajectory forecasting. The main contributions of this work are summarized as follows.

(1) We propose an MMFE module for calibrated multimodal representation learning, which explicitly models multimodal reliability and asymmetric cross-modal dependencies across modalities and temporal scales to construct semantically aligned motion representations.

(2) We introduce MEHP, a hierarchical prediction framework for future-motion multiscale refinement, which performs prediction-aware future-motion refinement through coarse uncertainty modeling, scale-aligned fusion, and progressive hierarchical correction.

(3) Extensive experiments and ablation studies on benchmark datasets validate the effectiveness of each proposed component and show consistent performance advantages compared with state-of-the-art trajectory prediction methods.

The remainder of this article is organized as follows. Section \uppercase\expandafter{\romannumeral 2} reviews related work on trajectory prediction and multiscale modeling. Section \uppercase\expandafter{\romannumeral 3} describes the proposed methodology. Section \uppercase\expandafter{\romannumeral 4} reports the experimental setup and results. Section \uppercase\expandafter{\romannumeral 5} concludes the paper.

\section{Related Works}

\subsection{Pedestrian Trajectory Prediction}

Understanding and predicting pedestrian trajectories and intentions are essential for safe interactions between humans and vehicles in autonomous driving \cite{ref116}. For trajectory prediction, early methods adopted physics-based models, and others, such as Gaussian processes \cite{ref11}, Bayesian methods \cite{ref12}, and Kalman filters \cite{ref13}, also achieved good performance. However, these approaches can only model uncertainty over short horizons and struggle with long sequences. In crowded scenes with diverse pedestrian motions, they are even less effective at handling the complexity.

Recent data-driven methods have become dominant in trajectory prediction due to their ability to extract rich features from complex environments. Early approaches employed Recurrent Neural Networks (RNNs) and Long Short-Term Memory (LSTM) models \cite{ref14, ref75} to capture sequential correlations in observed pedestrian trajectories and predict future paths. Generative Adversarial Networks (GANs) \cite{ref15,ref16} were later introduced to model pedestrian path distributions. To capture spatial interactions among agents, Graph Neural Networks (GNNs) \cite{ref2, ref60, ref61} have also been adopted, where pedestrians are represented as nodes and their interactions are encoded through graph structures.

Inspired by advances in Natural Language Processing (NLP), Transformer models \cite{ref64, ref79, ref106} have been widely applied to trajectory prediction for modeling long-range dependencies and global context. SNARTF \cite{ref65} uses a graph attention module to jointly model social and temporal interactions through sparse attention and identity cues, enabling efficient parallel prediction of diverse future trajectories. TUTR \cite{ref63} employs a transformer encoder-decoder to jointly capture global trajectories, social interactions, and multimodal futures, allowing parallel prediction without post-processing.

Diffusion models \cite{ref67, ref113, ref115}, a class of generative methods, model data generation as a gradual transformation from noise to the target distribution via a parameterized Markov chain, and have been explored to capture the stochastic nature of human motion. Guided diffusion is adopted in \cite{ref68} to generate controllable trajectories under motion constraints. MoFlow \cite{ref102} introduces a conditional flow-matching framework to balance trajectory accuracy and diversity, while IAD \cite{ref114} improves generation accuracy through refined diffusion priors and clean-manifold guidance. 

While these methods improve multimodal uncertainty modeling, they usually involve more complex generation processes with higher training and inference costs. In contrast, our framework emphasizes efficient uncertainty-aware coarse initialization for subsequent hierarchical refinement, which is better suited when efficiency and stable inference are important.

\subsection{Multimodal Data Based Trajectory Prediction }
Multimodal data plays an important role in motion prediction for autonomous driving, as it provides rich traffic cues and contextual information.  Several studies \cite{ref119, ref85, ref86, ref117, ref118} leverage multimodal inputs to improve pedestrian behavior understanding and trajectory prediction. UEN \cite{ref105} combines CNN-based scene features with a polar representation to model scene constraints and social interactions, capturing local cues for multimodal motion while reducing redundant predictions. Di-Long \cite{ref83} combines visual inputs with knowledge distillation, where a teacher with longer observations guides a student for long-term trajectory prediction to reduce uncertainty.

In the context of human motion prediction, cues such as human actions, characteristics, and pose are incorporated into motion forecasting. \cite{ref80} introduces an egocentric two-tower predictor that fuses multimodal inputs from both the vehicle and pedestrian sides, and employs an action-aware loss to decouple motions for improved flexibility and intention estimation. TSNet \cite{ref73} is a two-stream predictor that integrates trajectories with character cues such as action and appearance, using sparse character graphs to filter irrelevant information and refine trajectory embeddings. PCTP-AGFL \cite{ref120} proposed a progressive LSTM-based trajectory prediction framework with adaptive gating and contextual attention, while DPITRA \cite{ref122} introduced a multimodal model that jointly predicts pedestrian intention and trajectories using denoising attention and CVAE. Integrating trajectory and probability estimation within a multitask framework, PMTPN \cite{ref66} employs learnable motion queries with a multi-layer decoder and a multi-gate mixture-of-experts to handle action variability and improve prediction performance. However, most existing multimodal trajectory prediction methods mainly focus on utilizing and fusing diverse cues for prediction, while cross-modal inconsistency or interference is less explicitly addressed. In particular, the asymmetric roles and time-varying reliability of different modalities across motion scales and scenarios are not sufficiently modeled, which may limit the construction of semantically consistent multimodal representations.

\subsection{Multiscale Based Prediction Approach}
Multiscale modeling has demonstrated effectiveness in capturing hierarchical representations across vision and perception tasks, including visual detection \cite{ref92}, image classification \cite{ref93}, human pose estimation \cite{ref94}, and point cloud forecasting \cite{ref95}. In motion prediction, recent studies have incorporated multiscale modeling into prediction frameworks.

TPNMS \cite{ref96} introduces a temporal pyramid network to model pedestrian motion at multiple tempos, using squeeze and dilation modulation to build hierarchical temporal features with multi-level supervision. Leveraging the Gabor transform, MlgtNet \cite{ref97} constructs multiscale representations that integrate global and local temporal dependencies into a unified representation. In addition, several approaches decompose motion cues into multiple temporal scales via spectral decomposition \cite{ref100} or low-rank decomposition \cite{ref101}, enabling separation of coarse and fine motion patterns. Another line of work explores multiscale social interaction modeling. GroupNet \cite{ref98} learns pair-wise and group-wise relations through a trainable multiscale hypergraph, while MART \cite{ref99} extends this idea with a hypergraph relational transformer for adaptive multiscale social reasoning. Although recent multiscale methods, such as TPNMS and MlgtNet, have shown the usefulness of multiscale temporal modeling, their multiscale features are mainly derived from observed motion and then used for direct prediction. In contrast, MUSCLE-Net further uses the coarse future estimate to guide scale alignment and progressive correction, so that multiscale cues are adjusted according to the predicted future state. This provides a different use of multiscale information, extending it from observation-side encoding to prediction-conditioned future-motion refinement.

In summary, although existing trajectory prediction studies have explored multimodal perception and multiscale modeling, prior methods mainly focus either on multimodal cue utilization and fusion or on multiscale motion encoding. Consequently, how to (1) construct reliable and semantically consistent multimodal representations by calibrating modality reliability and mitigating cross-modal inconsistency, and (2) adaptively update predefined scale features according to predicted future motion scales, remains less explored.

\section{METHODOLOGY}

\subsection{Problem Definition}
Pedestrian trajectory prediction aims to estimate future pedestrian positions given observed motion history and contextual cues. We consider a scene containing multiple pedestrians, indexed by $i$. For each pedestrian, the model takes as input multimodal observations over an observed time horizon $t_{obs}$.

Specifically, pedestrian motion is represented by a sequence of 2D bounding boxes $B_{1:t_{obs}}^i = \{ B_t^i \in \mathbb{R}^4 \mid 1 \le t \le t_{obs} \}$, where $B_t^i = (x_t^{TL}, y_t^{TL}, x_t^{BR}, y_t^{BR})$ denotes the top-left and bottom-right coordinates of the bounding box at time $t$. In addition, the ego-vehicle motion is characterized by its speed sequence $V_{1:t_{obs}} = \{ V_t \in \mathbb{R} \mid 1 \le t \le t_{obs} \}$, and the pedestrian pose context is described by $P_{1:t_{obs}}^i = \{ P_t^i \in \mathbb{R}^{36} \mid 1 \le t \le t_{obs} \}$, where $P_t^i$ contains the 2D coordinates of 18 human body joints extracted using OpenPose. Given these multimodal observations, the goal is to predict the future pedestrian trajectory over a prediction horizon $t_{pred}$, denoted as $Y_{1:t_{pred}}^i = \{ Y_t^i \in \mathbb{R}^4 \mid 1 \le t \le t_{pred} \}$. The model outputs the predicted trajectory $\hat{Y}_{1:t_{pred}}^i = \{ \hat{Y}_t^i \in \mathbb{R}^4 \mid 1 \le t \le t_{pred} \}$, where $\hat{Y}_t^i = (x_t^{TL}, y_t^{TL}, x_t^{BR}, y_t^{BR})$ represents the estimated bounding box of the pedestrian at future time $t$.

\begin{figure*}
\centering
\includegraphics[scale= 0.73]{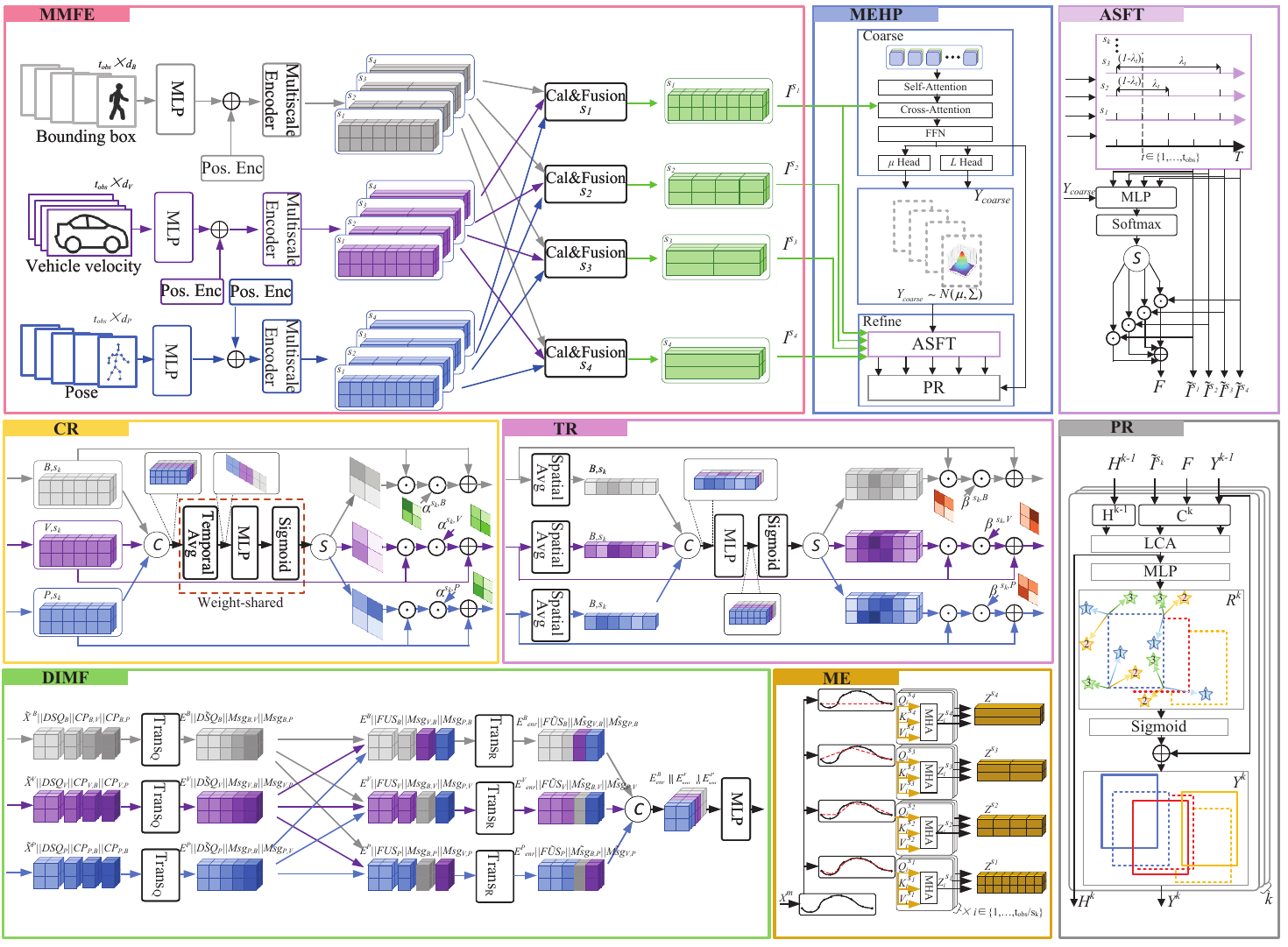}
\caption{The structure of the proposed MUSCLE-Net framework. The proposed framework consists of two core modules. The Multiscale Multimodal Feature Extraction Module (MMFE) integrates a multiscale encoder, channel- and temporal-wise recalibration, and directional multimodal interaction to construct unified and semantically aligned multiscale motion features. The Multiscale Enhanced Hierarchical Prediction Module (MEHP) then generates future trajectories through a Gaussian-based coarse predictor, an aligned scale fusion mechanism with temporal modulation, and a prediction-aware refinement stage, where predictions are iteratively corrected using multiscale contextual cues.}
\centering
\label{figure structure2}
\end{figure*}

\subsection{Approach Overview}
The overall framework of MUSCLE-Net is shown in Figure \ref{figure structure2}. Given multimodal observations, MMFE learns multiscale representations through multiscale modeling, feature recalibration, and directional fusion, which are then passed to MEHP for coarse prediction and refinement. Unlike prior methods that mainly use multiscale modeling for historical trajectory encoding or multimodal fusion for direct prediction, MUSCLE-Net addresses these two aspects through MMFE and MEHP. In particular, MMFE mainly operates at the representation level by constructing aligned representations across modalities and scales, whereas MEHP mainly operates at the prediction-refinement level by modeling future motion from coarse global trends to finer local details through hierarchical refinement.

\subsection{Multiscale Multimodal Feature Extraction Module}
Multimodal observations often exhibit scale-dependent motion patterns and time-varying reliability across modalities. To address these issues, we design the Multiscale Multimodal Feature Extraction (MMFE) module as the representation learner in our framework. Rather than serving as a generic multimodal fusion block, MMFE is designed to mitigate cross-modal inconsistency and enhance representation reliability before hierarchical prediction.

\subsubsection{Multiscale Encoder}

For each modality, we first project the raw input features into a shared latent space using a modality-specific MLP with positional encoding $\mathrm{PE}$, yielding $X^{m} = \mathrm{MLP}_{m}(X^{m}_{\text{raw}}) + \mathrm{PE}_{m} \in \mathbb{R}^{t_{\text{obs}} \times d}$, where $m \in \{B, V, P\}$ denotes bounding box, ego-vehicle velocity, and pedestrian pose, respectively. This projection unifies feature dimensions across modalities and injects temporal order information. We then adopt window sizes $s_k \in \{1, 3, 5, 15\}$ to partition $X^{m}$ along the temporal dimension into multiscale representations $X^{s_k,m} \in \mathbb{R}^{(t_{\text{obs}}/s_k) \times s_k \times d}$.

For modality $m$ and scale $s_k$, let $X^{s_k,m}_{i,j} \in \mathbb{R}^{d}$ denote the $j$-th feature in the $i$-th window, where $i \in \{1,\dots, t_{\text{obs}}/s_k\}$ and $j \in \{1,\dots,s_k\}$. The window mean is used as the query to provide a scale-consistent summary of local motion context, i.e., ${Q}^{\,s_k,m}_i= \frac{1}{s_k} \sum_{j=1}^{s_k} {X}^{\,s_k,m}_{i,j}\in \mathbb{R}^{d}$, while keys and values are obtained by projection as ${K}^{\,s_k,m}_{i}= f_{LN}\,({X}^{\,s_k,m}_{i}) \in \mathbb{R}^{s_k \times d}$ and ${V}^{\,s_k,m}_{i} = f_{LN}\,({X}^{\,s_k,m}_{i}) \in \mathbb{R}^{s_k \times d}$, where $f_{LN}$ denotes an MLP layer. Features within each window are then aggregated by multi-head attention, and stacking all windows yields ${Z}^{\,s_k,m}$:
\begin{equation}
{Z}^{\,s_k,m}_i
= \mathrm{MHA}\big(
    {Q}^{\,s_k,m}_i,
    {K}^{\,s_k,m}_i,
    {V}^{\,s_k,m}_i
  \big)
\in \mathbb{R}^{d},
\end{equation}
\begin{equation}
{Z}^{\,s_k,m}
= [{Z}^{\,s_k,m}_1 \Vert\ \dots \Vert\ {Z}^{\,s_k,m}_{t_{\text{obs}}/s_k}]
\in \mathbb{R}^{t_{\text{obs}}/s_k \times d}.
\end{equation}

Although these multiscale representations capture local motion tendencies, unimodal observations still suffer from sensing limitations, leading to uncertain estimates and inconsistent semantics. To mitigate these issues before multimodal fusion, we recalibrate modality-specific features to suppress unreliable cues and enhance semantic consistency. This design enables the subsequent fusion stage to better exploit modality-specific strengths across temporal scales.


\subsubsection{Channel-Wise Recalibration}
To harmonize heterogeneous modalities while preserving their distinctive characteristics, we introduce a Channel Recalibration (CR) module, drawing inspiration from \cite{ref110} and \cite{ref111}. Unlike modality-specific recalibration schemes, CR employs a shared gating mechanism to explicitly capture cross-modal channel dependencies and generate a unified channel-wise importance vector, which is subsequently applied to each modality.

We first cascade all three modality features into $Z^{s_k} = [ Z^{s_k, B}\Vert\ Z^{s_k, V}\Vert\ Z^{s_k, P} ] \in \mathbb{R}^{t_{obs}/{s_k} \times c}$, where $c = 3d$, and perform temporal average pooling to obtain a channel descriptor ${G^{s_k}}\in\mathbb{R}^{c}$, which summarizes each channel’s activation over time, enabling the model to reason about cross-modal channel dependencies. A lightweight fully connected layer $f_{LN}$ then produces a channel-wise calibration vector $W^{s_k}$:
\begin{equation}
\begin{aligned}
{G^{s_k}}=\frac{1}{L}\sum_{t=1}^{L}Z^{s_k}_{t},\,
{W^{s_k}}=\sigma\!(f_{LN}\,({G}^{s_k}))\in(0,1)^{c},
\end{aligned}
\end{equation}
where $L = t_{obs}/s_k$ denotes the number of temporal windows at scale $s_k$. $\sigma\!$ denotes the sigmoid function. For each modality $m$, we apply a learnable attenuation coefficient $ \alpha \in \mathbb{R}^{d}$, whose sigmoid output acts as a residual scaling factor:
\begin{equation}
\begin{aligned}
\tilde{{Z}}^{s_k,m} = {Z}^{s_k,m}
+ \sigma\!(\alpha^{s_k,m})\odot ({Z}^{s_k,m} \odot ({W}^{s_k,m}-{1})),
\end{aligned}
\end{equation}
where $W^{s_k,m} \in (0, 1)^{d}$ is the corresponding calibration vector for each modality, split from the calibration vector $W^{s_k}= [ W^{s_k, B}\Vert\ W^{s_k, V}\Vert\ W^{s_k, P} ]$.

\subsubsection{Temporal-Wise Recalibration}
After recalibrating globally unreliable channels, we further introduce a Temporal Recalibration (TR) module to handle frame-wise unreliability and asynchronous dynamics across modalities. At each time step $t$, we compress each modality to a compact scalar descriptor, then expand it to a channel-wise temporal gate by averaging over the channel dimension $d$, and a lightweight shared MLP $f_{LN}$ produces modality-wise gates $M^{s_k,m}_{t} \in {(0,1)^{d}}$:
\begin{equation}
\begin{aligned}
P^{s_k, m}_{t} = \frac{1}{d}\sum_{c=1}^{d} \tilde{Z}^{s_k, m}_{t,c}, \,   M^{s_k, m}_{t}=\sigma\!(f_{LN}({P}^{s_k, m}_{t})).
\end{aligned}
\end{equation}

For stability, each modality is equipped with a learnable per-channel attenuation vector $\beta \in \mathbb{R}^{t_{obs} \times d}$, whose sigmoid controls the update strength. The calibrated features are:
\begin{equation}
\begin{aligned}
\tilde{X}^{s_k,m}_{t}
= \tilde{Z}^{s_k,m}_{t}
+ \sigma\!\bigl(\beta^{s_k,m}_{t}\bigr)\odot
(\tilde{Z}^{s_k,m}_{t}\odot(M^{s_k,m}_{t}-1)).
\end{aligned}
\end{equation}

\subsubsection{Directional Interactive Multimodal Fusion}
To exploit complementary and asymmetric multimodal cues, inspired by \cite{ref112}, 
we propose a Directional Interactive Multimodal Fusion (DIMF) module that enables 
target-aware feature extraction and controlled integration through asymmetric 
information flow. Specifically, fusion is decomposed into a sender-side extraction 
stage and a receiver-side integration stage.

Given the calibrated feature sequence $\tilde{X}^{s_k, m}$ of each modality $m$ at scale $s_k$, we aim to extract distinct information streams $Msg_{m,n} \in \mathbb{R}^{1 \times d}$ for every target modality $n \neq m$. For each ordered pair $(m,n)$, we introduce two learnable tokens: a directional selection query $DSQ_m \in \mathbb{R}^{1 \times d}$, which conditions the sender modality $m$ and governs how information is extracted from its feature stream, and a directional communication prompt $CP_{m, n} \in \mathbb{R}^{1 \times d}$, which acts as a container that aggregates the distilled information specifically intended for modality $n$. A shared query transformer layer then processes the augmented sequence $Q^{s_k,m}$. This design contrasts with conventional global message passing and enables finer, modality-specific information routing.
\begin{equation}
\begin{aligned}
&\quad\quad{Q}^{s_k,m} = [\tilde{{X}}^{s_k, m} \Vert\ DSQ_{m} \Vert\ CP_{m,n_1} \Vert\ CP_{m,n_2} ],\\
&[{E}^{s_k, m}, \tilde{DSQ}_m, Msg_{m,n_1}, Msg_{m,n_2}]=\text{Qtrans}(Q^{s_k,m}),
\end{aligned}
\end{equation}
where ${E}^{s_k, m}$ is the updated intra-modal sequence, $Msg_{m, n_i}$ are tailored messages to targeted modalities $n_i$.

On the receiver side, each modality $n$ collects two directional messages from its sender modalities $m_i$ and, together with its fusion slot $FUS_n\in \mathbb{R}^{1 \times d}$, feeds them into a shared receiver transformer layer, enabling controlled integration of geometric cues, motion dynamics, and pose stability from different modalities.
\begin{equation}
\begin{aligned}
&\quad \quad R^{s_k,m}=[{E}^{s_k, n} \Vert\ FUS_{n} \Vert\ Msg_{m_1,n} \Vert\ Msg_{m_2,n} ],\\
&[{E}_{enr}^{s_k, n}, \tilde{FUS}_n, \tilde{Msg}_{m_1,n},\tilde{Msg}_{m_2,n}]=\text{Rtrans}(R^{s_k,m}),
\end{aligned}
\end{equation}
where ${E}_{enr}^{s_k, n}$ is enriched stream sequence, 

After directional enrichment, we concatenate the three enhanced streams and apply a lightweight MLP fusion layer $f_{LN}$ to obtain a unified feature representation, which is then passed to the subsequent prediction module.
\begin{equation}
\begin{aligned}
I^{s_k}
= f_{LN}(
E_{enr}^{s_k, B} \Vert
E_{enr}^{s_k, V} \Vert
E_{enr}^{s_k, P}
)
\in \mathbb{R}^{L \times d},
\end{aligned}
\end{equation}
where $L = t_{obs}/s_k$ denotes the number of temporal windows at scale $s_k$.

DIMF allows each modality to contribute differently depending on the prediction target. For example, pose cues are more informative for local motion refinement, whereas bounding-box dynamics dominate long-term trend estimation. This helps reduce interference from noisy modalities and provides scale-aware cues for future motion refinement.

\subsection{Multiscale Enhanced Hierarchical Prediction Module}
Existing multiscale prediction modules often operate directly on fused features, which makes the multiscale characteristics of future motion less explicitly modeled. To address this issue, we introduce the Multiscale-Enhanced Hierarchical Prediction (MEHP) module, a three-stage coarse-to-fine pipeline for trajectory forecasting. Rather than serving as a generic multistage decoder, MEHP is designed to model future motion from coarse global structure to finer local details.

\subsubsection{Coarse Prediction}

In the coarse stage, we use only the finest-scale feature $\tilde{X}^{s_{k=1}}$ to obtain the coarse representation $H_{coarse}$ and coarse bounding boxes $\hat{Y}_{coarse}$. We model each per-frame bounding box (with $D=4$) using a multivariate Gaussian distribution, $\hat{Y}^{\,i}_{coarse,t}\sim \mathcal{N}\!\bigl(\mu^{i}_{t},\,\Sigma^{i}_{t}\bigr)$, whose probability density is defined as:
\begin{equation}
p(Y;\mu,\Sigma)
= \frac{1}{(2\pi)^{D/2}\,|\Sigma|^{1/2}}
\exp\! (-\tfrac{1}{2}\,(Y-\mu)^{\top}\Sigma^{-1}(Y-\mu)).
\end{equation}

For convenience, we parameterize the covariance via the Cholesky factorization $\Sigma=LL^\top$ and rewrite the density:
\begin{equation}
p(Y;\mu,L)
= \frac{1}{(2\pi)^{D/2}\,|L|}
\,\exp\! (-\tfrac{1}{2}\bigl\|L^{-1}(Y-\mu)\bigr\|_{2}^{2}),
\end{equation}
where $\mu^{i}_{t}=[x_1,y_1,x_2,y_2]\in\mathbb{R}^4$ and $L^{i}_{t}\in\mathbb{R}^{D\times D}$ is the lower-triangular Cholesky factor with positive diagonal. $|\Sigma|^{1/2}=|L|$. The coarse module follows a standard Transformer block with self-attention, feed-forward layers, and normalization to produce coarse embeddings:
\[
H_{\text{coarse}}=\mathrm{Transformer}\bigl(I^{s_{k=1}}\bigr).
\]

Two MLP heads then predict the mean $\mu \in \mathbb{R}^{t_{pred} \times 4}$ and the Cholesky factor $L \in \mathbb{R}^{t_{pred} \times D \times D}$, respectively. During training, coarse trajectory samples are drawn from the predicted Gaussian distribution $\mathcal{N}(\mu, LL^\top)$ using the reparameterization formulation. During inference, however, the deterministic trajectory estimate is obtained using the mean prediction $\mu$ of the Gaussian distribution as the coarse prediction.

\subsubsection{ASFT Fusion}
Aligned Scale Fusion with Temporal Modulation (ASFT) aims to align multiscale features and fuse them according to the scale characteristics of future motion. To enable per-frame multiscale fusion, we align multiscale features $I^{s_k}$ to the original temporal resolution $t_{obs}$ using per-channel linear interpolation. For frame index $t\in \{1,...t_{obs}  \}$, define the aligned temporal index at each scale $s_k$:
\begin{equation}
\begin{aligned}
u_t^{s_k} = t/s_k,\
i_0=\bigl\lfloor u_t^{s_k} \bigr\rfloor, \
i_1=\min\!\bigl(i_0+1,\,L_{s_k}-1\bigr),
\end{aligned}
\end{equation}
where $\lfloor\cdot\rfloor$ denotes the floor operator, $\min(\cdot)$ clamps the index to the last valid position, and $L_{s_k}= t_{obs}/s_k$. Then the aligned feature at frame $t$ is calculated through linear interpolation: 
\begin{equation}
\begin{aligned}
\tilde{I}_t^{s_k} = (1-\lambda_t)\,{I}^{s_k}_{i_0} + \lambda_t\,{I}^{s_k}_{i_1}
\in \mathbb{R}^{d},
\end{aligned}
\end{equation}
where $\lambda_t = u_t^{s_k}-i_0$ is the fractional interpolation weight for frame $t$ after mapping to the downsampled timeline.

To guide the model in selecting appropriate scale features according to bounding-box size, the area feature is included and embedded ${I}_t^{\text{area}} = f_{LN}\!\bigl(\log(1+a_t)\bigr)\in\mathbb{R}$, where $a_t$ is the coarse box area based on $\hat{Y}_{\text{coarse}}$ and $f_{LN}$ is an MLP. Finally, we fuse the aligned features per frame by a convex combination, yielding the fused tube $F$ for downstream refinement:
\begin{equation}
\begin{aligned}
{C}_t
= \bigl[\, \tilde{I}_{t}^{s_1},\, \tilde{I}_{t}^{s_2},\, \cdots,\, \tilde{I}_{t}^{s_{k}},\, {I}_t^{\text{area}} \,\bigr]
\in \mathbb{R}^{\,k \times (d+1)},
\end{aligned}
\end{equation}
\begin{equation}
\begin{aligned}
F_t = \sum_{k\in K} w_{t,s_k}\,\tilde{Z}_{t}^{s_k}\in \mathbb{R}^{d},\ {w}_t = \text{softmax}(f_{LN}(C_t)) \in \mathbb{R}^k.
\end{aligned}
\end{equation}

\subsubsection{Prediction-Aware Refinement} 
After obtaining coarse bounding boxes $\hat{Y}_{\text{coarse}}$, prediction features $H_{\text{coarse}}$, and aligned multiscale features $\{\tilde{I}^{s_k}\}_{k=1}^{K}$, we introduce a prediction-aware refinement module that uses coarse future cues to guide feature-aware trajectory correction from global structure to finer local details.

Specifically, at each refinement scale $s_k$, we refine the input box $Y^{k-1}$ by aggregating multiple candidate corrections predicted from the current-scale features. We first compute an area embedding from $Y^{k-1}$: $I^{area,k} = f_{LN}^{k}\!\bigl(\log(1+a^{k-1})\bigr)\in\mathbb{R}$, where $a^{k-1}$ denotes the coarse bounding-box area derived from $Y^{k-1}$ and $f_{LN}^{k}$ is an MLP. Given query features $H^{k-1}$ from the previous refinement stage and key/value features constructed from the context $C^k = [\hat{I}^{s_k}, F, I_{\text{area},k}]$ at scale $s_k$, the localized cross-attention uses the current coarse future state to retrieve locally relevant correction cues and produces $N$ candidate correction anchors $R^k \in \mathbb{R}^{N \times 4}$:
\begin{equation}
\begin{aligned}
\{\, R_{n}^{k}\,\}_{n=1}^{N}
= f_{\mathrm{LCA}}^{k}\! (H^{\,k-1},\, C^{\,k}),
\end{aligned}
\end{equation}
where $f^k_{LCA}$ denotes localized cross-attention. For a temporal window radius $i$, we build a context set $N(t)=\{t-i,\dots,t+i\}$ with boundary padding at $t=1$ and $t=t_{pred}$. The stacked features within this window, ${C}_{N(t)}=\bigl[C_{t-i},\dots,C_{t+i}\bigr]\in\mathbb{R}^{(2i+1)\times d_{kv}}$, are used as attention keys and values.

To aggregate these candidates, we compute distance-decay weights $w^{k} \in \mathbb{R}^{t_{pred} \times N \times 4}$ in the center space of the previous refined box \(Y^{k-1}\), so that proposals closer to the center are emphasized, while far-away, implausible corrections are smoothly suppressed:
\begin{equation}
\begin{aligned}
&\quad \quad \quad \tilde{w}_{t,n}^{k}
= \frac{w_{t,n}^{k}}{\sum_{m=1}^{N} w_{t,m}^{k}},\\
&w_{t,n}^{k}
= \frac{1}{\,1+\alpha\,\bigl\lVert c(R_{t,n}^{k})-c(Y_{t}^{\,k-1})\bigr\rVert^{\beta}},
\end{aligned}
\end{equation}
where $\tilde{w}^{k} \in \mathbb{R}^{t_{pred} \times N \times 4}$ denotes the normalized weights. The hyperparameters are set to $\alpha=1.0$ and $\beta=2.0$. $c(\cdot)$ maps a bounding box to its center. We then form a gated residual and update the box in the current refinement stage:
\begin{equation}
\begin{aligned}
&\Delta Y_t^{\,k}
= \sum_{n=1}^{N} \tilde{w}_{t,n}^{\,k}\,(R_{t,n}^{\,k}-Y_t^{\,k-1}), \\
&Y_t^{\,k}=Y_t^{\,k-1}+\Gamma^{k} \odot\Delta Y_t^{\,k},
\end{aligned}
\end{equation}
where $\Gamma^{k}=\sigma({g^k})\in(0,1)^{4}$ is learned per-coordinate gate that stabilizes fine-grained corrections and $g^k \in \mathbb{R}^4$ is learnable parameter vector.

This stage performs prediction-aware local correction by using coarse future estimates to dynamically select scale-relevant contextual cues, thereby transforming coarse high-level predictions into spatially precise bounding boxes.

\subsection{Training Optimization}
The overall loss has two components, a coarse stage and a refinement stage. In the coarse stage, we encourage the coarse head to be both diverse and well-calibrated. Diversity encourages coverage of multiple plausible futures, while calibration ensures the predicted uncertainty faithfully reflects trajectory dispersion. To promote diversity, we use a Best-of-M sampling scheme where $M$ candidates are drawn from the predicted Gaussian with the reparameterization trick:
\begin{equation}
\begin{aligned}
\tilde{Y}^{m}_{t}= {\mu}_{t} + L_{t}\,\odot \, \epsilon^{m}_{t},\qquad \epsilon^{m}_{t}\sim\mathcal N( 0, I) \in \mathbb{R}^{D},
\end{aligned}
\end{equation}
where $m \in \{1,\dots,M\}$ indexes the sampled candidates. $L$ is the Cholesky factor.

All sampled candidates are scored by a sequence-level $L_1$ distance. Only the candidate $m^{*}$ with the smallest sequence distance to the ground truth is used for backpropagation, and the loss for keeping diversity is given: 
\begin{equation}
\begin{aligned}
&\qquad Dis^{m} = \sum_{t}\lVert\tilde{Y}_{t}^{m}-Y_{t}\lVert_1, \\
\!\!&\mathcal{L}_{\mathrm{div}}= Dis^{m^*}, \quad m^{*} = \operatorname*{arg\,min}_{m} Dis^{m},
\end{aligned}
\end{equation}
where $Y$ is the ground truth trajectory.

To ensure calibration, the Gaussian negative log-likelihood is used to optimize the distribution:
\begin{equation}
\mathcal{L}_{\mathrm{NLL}}(Y;\mu,L) = \tfrac{1}{2}\!\Bigl[\bigl\|L^{-1}(Y-\mu)\bigr\|_2^2\;+\; 2\sum_{i=1}^{D}\log L_{ii}\Bigr],
\end{equation}
\begin{equation}
\mathcal{L}_{\mathrm{cal}}
=\frac{1}{N \times t_{pred}}\sum_{n=1}^{N}\sum_{t=1}^{t_{pred}}
\mathcal{L}_{\mathrm{NLL}}\!\bigl(Y_{n,t};\,\mu_{n,t},\,L_{n,t}\bigr),
\end{equation}
where $N$ is the total number of samples.

For the final refinement stage, we directly supervise the refined trajectory using an $L_1$ loss. Finally, the overall objective combines the refinement loss with the coarse-stage terms:
\begin{equation}
\begin{aligned}
\mathcal{L}_{\mathrm{ref}}
= \frac{1}{N \times t_{pred}}\sum^{N}_{n}\sum^{t_{pred}}_{t=1}
\bigl\lVert {Y}_{n,t}^{\mathrm{final}} - {Y}_{n,t} \bigr\rVert_{1}\,,
\end{aligned}
\end{equation}
\begin{equation}
\begin{aligned}
\mathcal{L}
= \mathcal{L}_{\mathrm{ref}}
+ \lambda_{\mathrm{coarse}}\!(
    \mathcal{L}_{\mathrm{div}}
    + \lambda_{\mathrm{nll}}\,\mathcal{L}_{\mathrm{cal}})\,,
\end{aligned}
\end{equation}
where we set $\lambda_{\mathrm{coarse}}=1.0$ and $\lambda_{\mathrm{nll}}=0.1$ empirically.

\section{Experiments and Results Analysis}

\subsection{Dataset}
Our experiments are conducted on two public benchmark datasets for pedestrian trajectory prediction: JAAD (Joint Attention for Autonomous Driving) \cite{ref71} and PIE (Pedestrian Intention Estimation) \cite{ref72}. JAAD consists of on-board video recordings from real driving scenarios, with hundreds of short sequences extracted from over 240 hours of footage. In contrast, PIE captures continuous urban traffic scenes and provides a comprehensive view of real-world dynamics. Both datasets offer synchronized ego-vehicle information, including speed, heading direction, and GPS signals, collected from onboard diagnostic sensors alongside video data. Following prior work, we use 15 observed frames (0.5 s) as input and predict pedestrian trajectories over future horizons of 15 frames (0.5 s), 30 frames (1.0 s), and 45 frames (1.5 s).

\subsection{Metrics}
To evaluate MUSCLE-Net, we adopt standard metrics commonly used in prior work. Mean Squared Error (MSE) measures the average squared error between predicted and ground-truth bounding boxes over the entire prediction horizon, computed on the top-left and bottom-right coordinates. Center MSE ($C_{\text{MSE}}$) reports the average squared distance between the predicted and ground-truth bounding box centers across all frames, while Center Final MSE ($CF_{\text{MSE}}$) measures the squared center-point error at the final prediction frame. All bounding box errors are computed using the upper-left and lower-right corner coordinates. During inference, the mean of the predicted Gaussian distribution is used as the deterministic coarse trajectory estimate for subsequent refinement. To maintain consistency with prior trajectory prediction works that adopt the best-of-N evaluation protocol, we run the model 20 times and report the minimum error among the predicted trajectories (best-of-20).

\subsection{Implementation}
Following prior work, all input sequences are normalized to mitigate the sensitivity of prediction performance to input scale. The bounding-box and center coordinates are transformed into a relative coordinate system by subtracting the coordinates of the first observed frame in each sequence, which reduces global translation variations. No additional trajectory smoothing or filtering is applied during preprocessing. All input modalities are projected into a 64-dimensional embedding space to ensure dimensional consistency. In the ME module, the number of attention heads is set to 4. Within the DIMF module, the $\mathrm{Trans}_{Q}$ and $\mathrm{Trans}_{R}$ layers share the same network architecture. For the MEHP module, the number of coarse sampling components is set to $M=4$, and the number of refinement anchors in PR is fixed to $N=5$. The distance-decay weighting parameters are set to $\alpha=1.0$ and $\beta=2.0$. The localized temporal attention window radius is fixed to $i=1$. The model is implemented in PyTorch and trained using the Adam optimizer. The learning rate is set to $1\times10^{-3}$, the batch size is 32, and a dropout rate of 0.1 is applied during training. All experiments are conducted on NVIDIA RTX 5090 GPUs with Intel Xeon 8481C CPUs. 

\subsection{Quantitative Evaluation}

\begin{table*}[!ht]
\centering
\renewcommand{\arraystretch}{0.9}
\setlength{\tabcolsep}{5pt}  
\fontsize{10}{10}\selectfont
\caption{ Quantitative Results (Pixels) on the JAAD and PIE Test Set. \textbf{Bold} is best, and \underline{underline} is suboptimal. All errors are reported in pixels.}
\label{result_1}
\begin{tabular}{l |c | c c c |c| c| c c c| c |c}
\toprule
\multirow{2}{*}{Method } & 
\multirow{2}{*}{Year} & 
\multicolumn{5}{c|}{JAAD} &
\multicolumn{5}{c}{PIE} \\
\cmidrule(lr){3-7} \cmidrule(lr){8-12}
& & 
\multicolumn{3}{c|}{MSE$\downarrow$} & 
$C_{\text{MSE}}$ $\downarrow$ & 
$CF_{\text{MSE}}$ $\downarrow$ & 
\multicolumn{3}{c|}{MSE$\downarrow$} & 
$C_{\text{MSE}}$ $\downarrow$ & 
$CF_{\text{MSE}}$ $\downarrow$ \\
\cmidrule(lr){3-5}\cmidrule(lr){6-7}\cmidrule(lr){8-10}\cmidrule(lr){11-12}
& & 0.5s & 1.0s & 1.5s & 1.5s & 1.5s & 0.5s & 1.0s & 1.5s & 1.5s & 1.5s \\
\midrule
PIE$_{traj}$ \cite{ref71}& 2019 & 110 & 399 & 1248 & 1183 & 4780 & 58 & 200 & 636 & 596 & 2477 \\
BiTraP-NP \cite{ref78} & 2021 & 38 & 94 & 222 & 177 & 565 & 23 & 48 & 102 & 81 & 261 \\
CTGF \cite{ref77}      & 2022 & 38 & 90 & 210 & 160 & 582 & 20 & 43 & 92  & 67 & 203 \\
ABC+ \cite{ref76}      & 2022 & 40 & 89 & 189 & 145 & 409 & 16 & 38 & 87  & 65 & 203 \\
SGNet-ED \cite{ref75}  & 2022 & 37 & 86 & 197 & 149 & 443 & 22 & 36 & 89  & 66 & 206 \\
AD-Sampler \cite{ref109}        & 2024 & 42 & 88 & 175 & 127 & 362 & 16 & 39 & 77  & 64 & 153 \\
ENCORE-NR \cite{ref74} & 2024 & \underline{32} & 85 & 210 & 162 & 554 & 15 & 33 & 70  & 49 & 155 \\
TSNet \cite{ref73}     & 2024 & 41 & 84 & 166 & 121 & 325 & 15 & 34 & 73  & 51 & 133 \\
DPITRA \cite{ref122}   & 2025 & 34 & 199 & 789 & 756 & 3121 & \textbf{8} & 71 & 225  & 201 & 1132 \\
PMTPN-T \cite{ref66}   & 2025 & 34 & \underline{71} & \underline{146} & \underline{105} & \underline{279} & \underline{14} & \underline{31} & \underline{67}  & \underline{47} & 131 \\
PMTPN-PT \cite{ref66}  & 2025 & 35 & 72 & 158 & 116 & 320 & 16 & 33 & 69  & 48 & \underline{128} \\
\midrule
MUSCLE-Net   & -    & \textbf{29} & \textbf{65} & \textbf{130} & \textbf{101} & \textbf{242} & \underline{14} & \textbf{28} & \textbf{61}  & \textbf{40} & \textbf{119} \\
\bottomrule
\end{tabular}
\end{table*}

We compare our proposed method with state-of-the-art first-person view trajectory prediction models, including DPITRA \cite{ref122}, PMTPN \cite{ref66}, TSNet \cite{ref73}, ENCORE-NR \cite{ref74}, AD-Sampler \cite{ref109}, SGNet-ED \cite{ref75}, ABC+ \cite{ref76}, CTGF \cite{ref77}, BiTraP-NP \cite{ref78}, and PIE\textsubscript{traj} \cite{ref71}. The results are summarized in Table \ref{result_1}.

The comparison with state-of-the-art methods shows that MUSCLE-Net achieves the best or second-best performance across most metrics on both the JAAD and PIE datasets. Specifically, for short-horizon prediction at 0.5 s and 1.0 s, our method reduces MSE relative to PMTPN-PT by 17.1\% and 9.7\% on JAAD, and by 12.5\% and 15.2\% on PIE, respectively, demonstrating clear advantages in early forecasting. For long-horizon prediction at 1.5 s, our approach further yields consistent improvements. On JAAD, MSE, $C_{\text{MSE}}$, and $CF_{\text{MSE}}$ are reduced by 17.7\%, 12.9\%, and 24.4\%, respectively, while on PIE the corresponding reductions reach 11.6\%, 16.7\%, and 7.0\%. The performance gains stem from the proposed multiscale feature modeling, which allows the network to adaptively exploit both fine-grained and coarse temporal cues under diverse motion patterns. As a result, the model effectively captures rapid motion variations while maintaining long-term trajectory consistency, leading to more stable and accurate predictions across different forecasting horizons.

\subsection{Qualitative Analysis}

\begin{figure*}[t]
    \centering
    \includegraphics[width=0.95\textwidth]{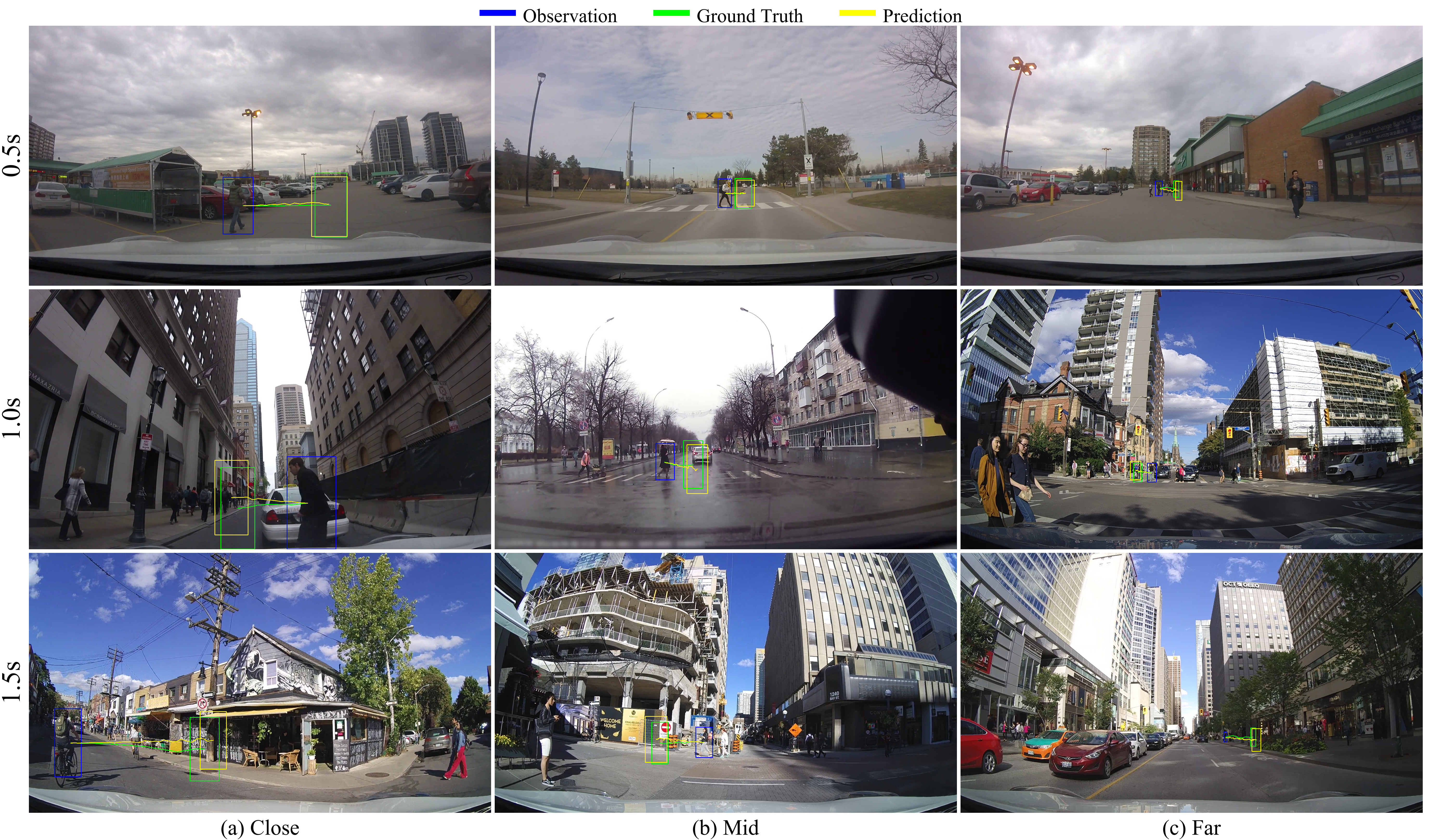}
    \caption{Visualization of the prediction results under various scenarios. The first to the third rows correspond to prediction horizons of 0.5 s, 1.0 s, and 1.5 s, respectively, while cases (a), (b), and (c) represent Close, Mid, and Far observation distances. The blue bounding box marks the last observed frame, and the green and yellow bounding boxes indicate the ground-truth and predicted boxes of the final future frame. The green and yellow lines further illustrate the ground-truth and predicted center-point trajectories over the prediction horizon.}
    \label{vis_1}
\end{figure*}
\begin{figure*}[!ht]
    \centering
    \includegraphics[width=0.90\textwidth]{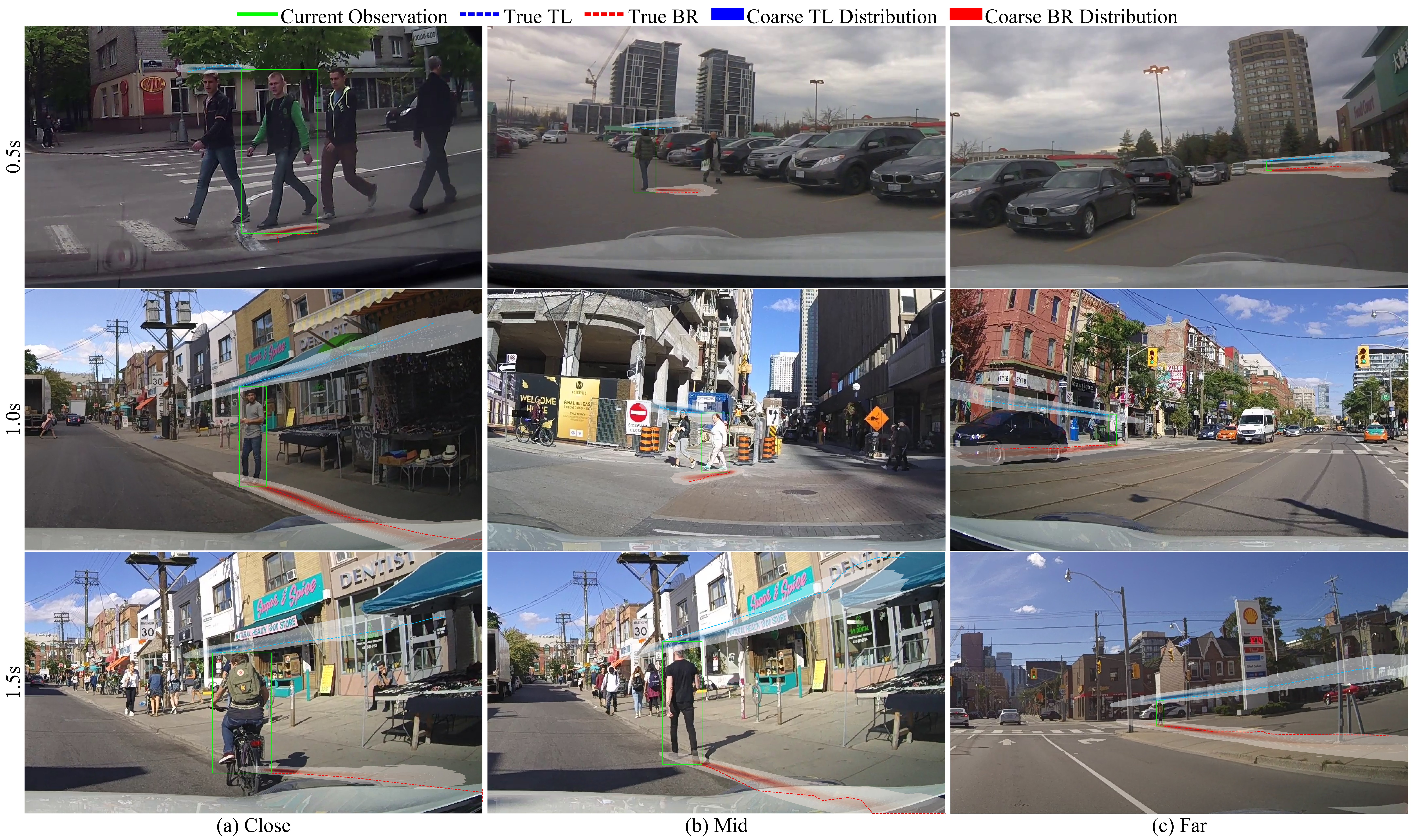}
    \caption{
Visualization of coarse prediction distributions at different time horizons and viewing distances.
From top to bottom, the rows correspond to prediction horizons of 0.5 s, 1.0 s, and 1.5 s, while the columns represent Close, Mid, and Far scenarios. In each subfigure, the green bounding box denotes the current observation, and the dashed blue and red boxes indicate the ground truth top left (TL) and bottom right (BR) corners. The blue and red shaded regions depict the coarse probability density distributions of the predicted top left and bottom right corner points across future frames.}
    \label{vis_dis_coarse}
\end{figure*}

\begin{figure}[!ht]\centering
	\includegraphics[width=8.4cm]{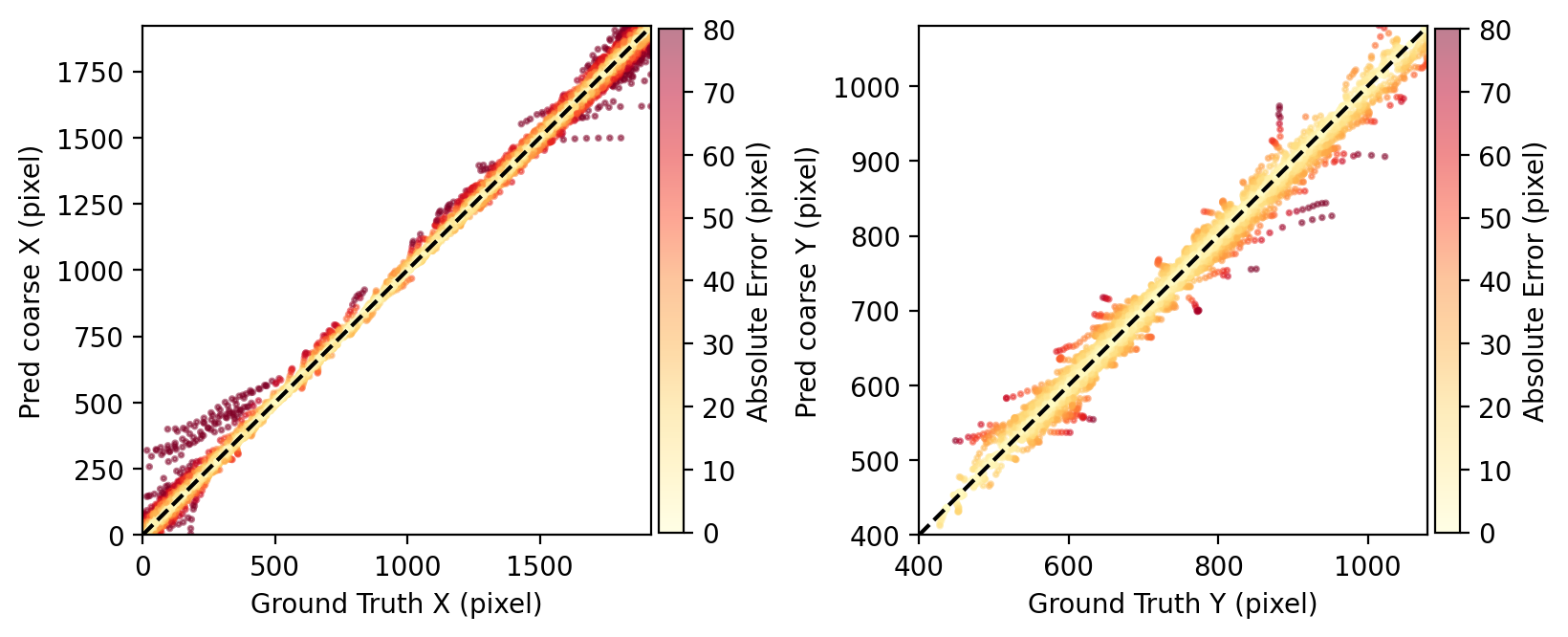}
    \includegraphics[width=8.4cm]{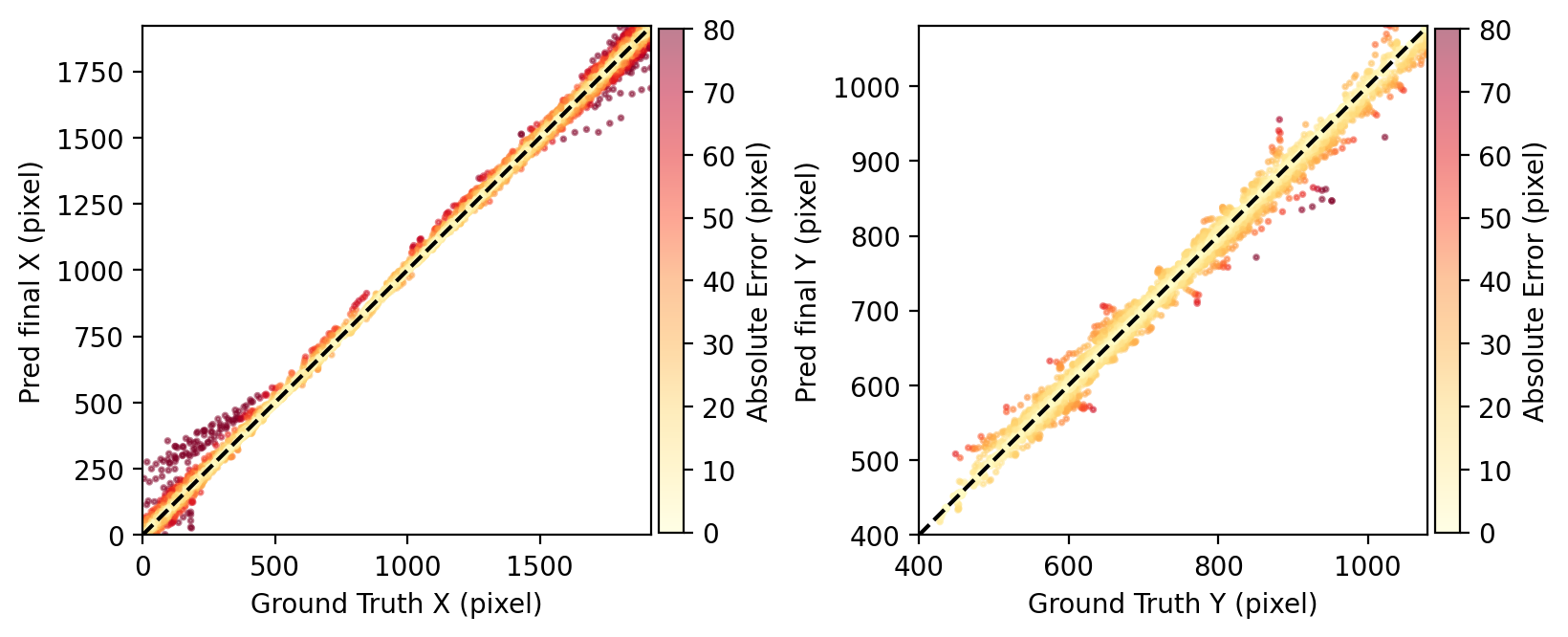}
	\caption{Illustration of the effectiveness of PR on the JAAD dataset. The top row shows the error distribution between the coarse predictions and the ground truth before applying PR, while the bottom row presents the error distribution between the final predictions and the ground truth after applying PR.
    }
    \label{vis_error_JAAD}
\end{figure}

\begin{figure}[!ht]\centering
	\includegraphics[width=8.4cm]{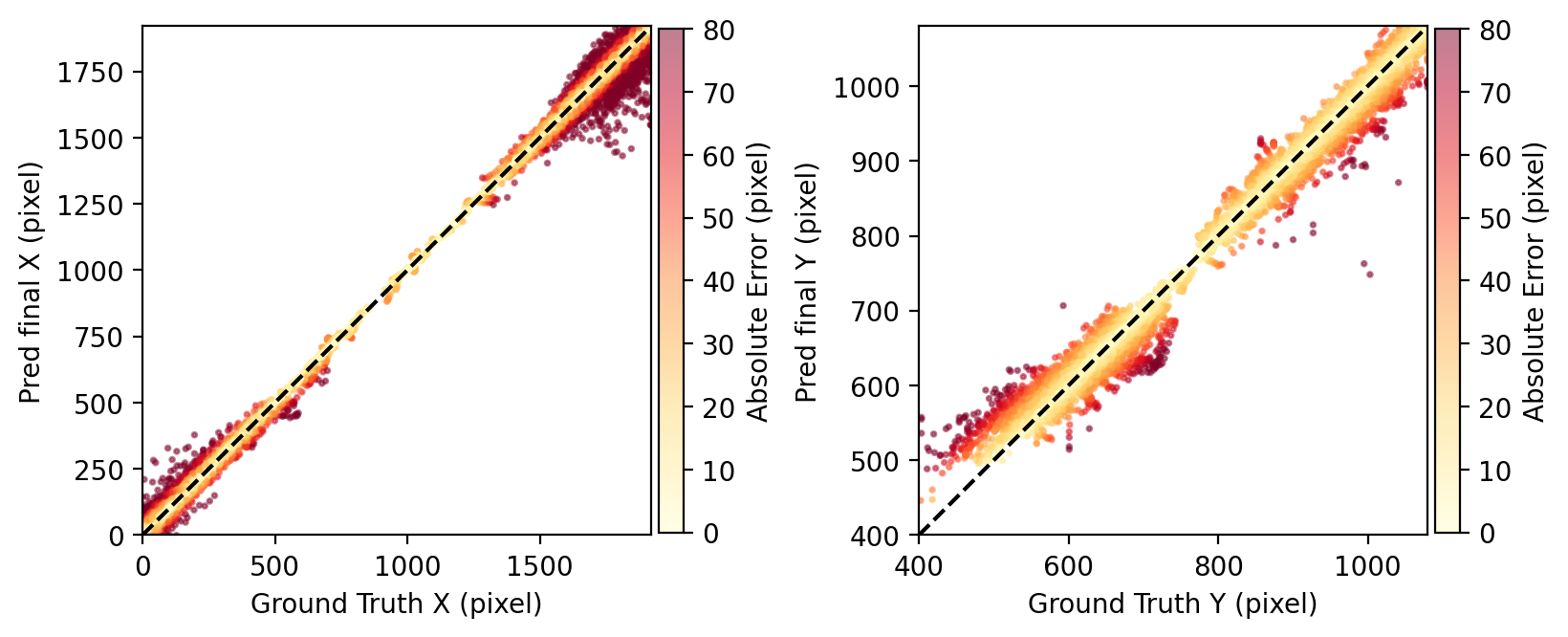}
    \includegraphics[width=8.4cm]{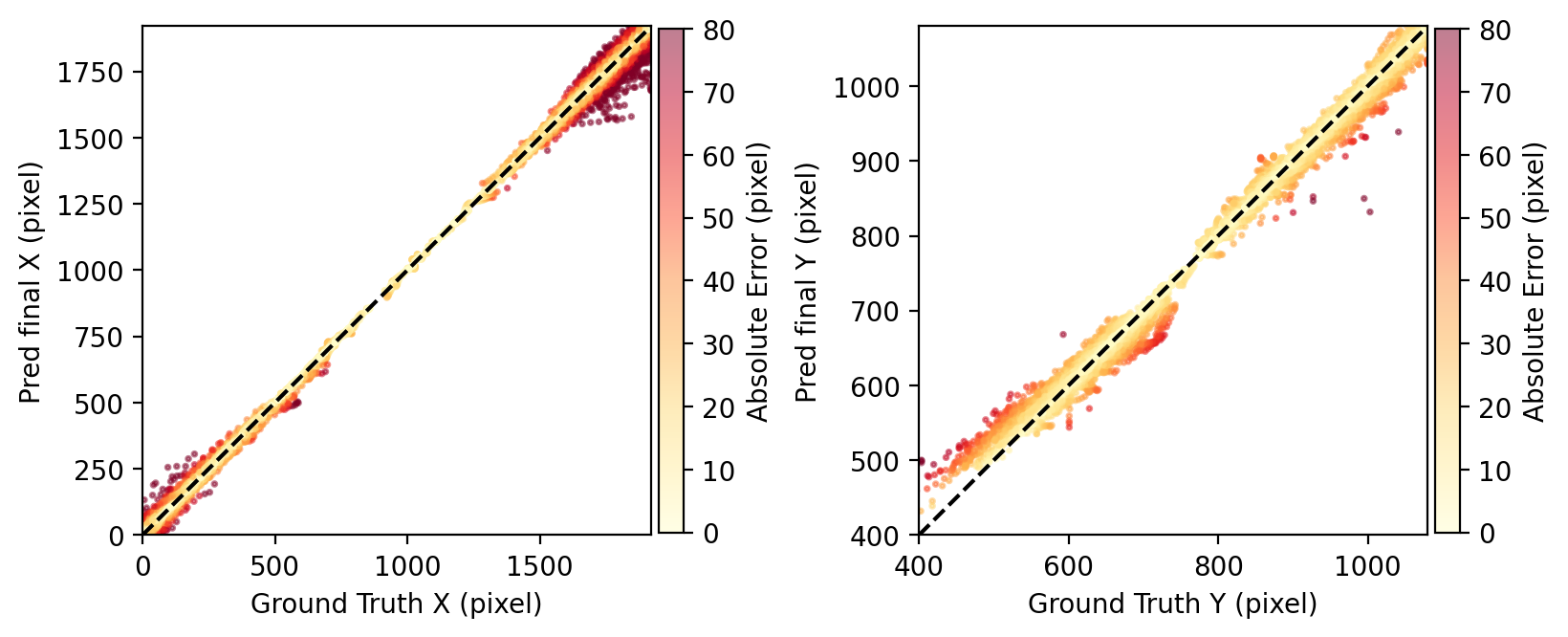}
	\caption{Illustration of the effectiveness of PR on the PIE dataset. The top row shows the error distribution between the coarse predictions and the ground truth before applying PR, while the bottom row presents the error distribution between the final predictions and the ground truth after applying PR. }
    \label{vis_error_PIE}
\end{figure}

\subsubsection{Trajectory Visualization}
The qualitative results in Figure~\ref{vis_1} demonstrate that MUSCLE-Net produces accurate and stable trajectory predictions across different viewing distances, including Close, Mid, and Far, as well as multiple prediction horizons of 0.5 s, 1.0 s, and 1.5 s. These scenarios involve substantial variations in bounding-box scale, visual appearance, and motion patterns, thereby providing a comprehensive evaluation of the model’s robustness.

In close scenes, large bounding boxes and rapid apparent motion amplify the impact of small localization errors. Even under such challenging conditions, our method delivers accurate multi-horizon predictions, with close alignment between predicted trajectories and ground truth. In mid-range scenes, predictions remain smooth and stable, successfully tracking pedestrians through directional changes and reflecting robust temporal modeling. In far-range scenes, despite diminished visual cues and small object scale, the proposed method preserves reliable motion trends, highlighting the generalization ability of the multiscale fusion and refinement design.

Across all distances and prediction horizons, the results demonstrate that the proposed approach yields coherent and accurate future bounding-box predictions, effectively addressing multiscale challenges, diverse pedestrian behaviors, and complex urban environments.

\subsubsection{Coarse Prediction Visualization}
Figure~\ref{vis_dis_coarse} visualizes the coarse-stage prediction distributions. The blue and red density maps correspond to the predicted top-left and bottom-right bounding box corners, respectively, providing a spatial depiction of predictive uncertainty over future time steps. Across close, mid, and far scenarios, the distributions exhibit coherent spatial patterns aligned with expected pedestrian motion. As the prediction horizon increases, the distributions become progressively more dispersed, reflecting the growing uncertainty in long-term forecasting and the increased variability of coarse bounding box estimates before refinement.

\subsubsection{Refinement Process Visualization}
Figures~\ref{vis_error_JAAD} and~\ref{vis_error_PIE} visualize the prediction errors before and after the refinement stage on the JAAD and PIE datasets, respectively. In the coarse predictions shown in the first row, the estimated bounding boxes generally follow the ground truth but exhibit deviations, particularly at larger coordinate values. Warmer colors indicate higher errors, suggesting that the coarse stage captures only a broad distribution of plausible future locations and suffers from elevated uncertainty. After refinement, shown in the second row, the predicted points align more closely with the ideal diagonal, with reduced scatter and consistently lower error across the spatial domain. This indicates that the refinement stage effectively corrects systematic biases inherited from the coarse predictions and yields more accurate and stable bounding box localization.

\subsection{Ablation Study}
In this section, ablation studies are conducted on the PIE and JAAD datasets to assess the contribution of each model component.

\subsubsection{Impact of Key Design Components}
In this subsection, we perform an extensive ablation study to systematically evaluate the contribution of each key component in the proposed framework, including Channel Recalibration (CR), Temporal Recalibration (TR), Directional Interactive Multimodal Fusion (DIMF), Aligned Scale Fusion with Temporal Modulation (ASFT), and Prediction-Aware Refinement (PR).

For the variant without ASFT, multiscale features are fused by weighted aggregation without explicit temporal alignment. Without PR, all scale-aligned features produced by ASFT are concatenated once and directly used to predict correction anchors, instead of being refined progressively. For the variant without DIMF, multimodal features are simply concatenated and passed through an MLP, removing directional interaction.

As shown in Table~\ref{ablation_components}, each component contributes to the overall performance. Removing CR or TR results in noticeable degradation, confirming the importance of channel-level stabilization and temporal reliability modeling. Replacing DIMF with simple concatenation leads to a performance drop, highlighting the necessity of directional information exchange rather than uniform fusion. Eliminating ASFT also degrades accuracy, demonstrating the benefit of temporally aligned multiscale features over naive aggregation. Finally, removing PR yields less precise bounding boxes, indicating that prediction-aware refinement is important for correcting geometric deviations. These results suggest that MMFE mainly improves multimodal representation quality, whereas MEHP mainly enhances future-motion refinement, rather than acting as redundant stacked components.

We further report the ablation results of the two high-level components in Tables~\ref{Ablation_keycomp1} and~\ref{Ablation_keycomp2}. For the MMFE-only setting, the prediction module directly concatenates multiscale features and uses an MLP for trajectory prediction. For the MEHP-only setting, only the multiscale representation is retained to obtain features at different scales. As shown in the two tables, removing either MMFE or MEHP leads to clear performance degradation on both JAAD and PIE, while the full model consistently achieves the best results. These results further suggest that MMFE and MEHP contribute different yet complementary modeling roles, with MMFE improving multimodal representation quality and MEHP enhancing future-motion refinement.

\begin{table*}[!ht]
\centering
\renewcommand{\arraystretch}{1.0}
\setlength{\tabcolsep}{6pt}  
\fontsize{10}{10}\selectfont
\caption{Ablation study of each component on the JAAD and PIE test sets. \textbf{Bold} is best. Lower is better.}
\label{ablation_components}
\begin{tabular}{
  >{\centering\arraybackslash}p{0.8cm}
  >{\centering\arraybackslash}p{0.8cm}
  >{\centering\arraybackslash}p{0.8cm}
  >{\centering\arraybackslash}p{0.8cm}
  >{\centering\arraybackslash}p{0.8cm}
  | c c c | c c | c c c | c c}
\toprule
\multicolumn{5}{c|}{Components} &
\multicolumn{5}{c|}{JAAD} &
\multicolumn{5}{c}{PIE} \\
\cmidrule(lr){1-5}\cmidrule(lr){6-10}\cmidrule(lr){11-15}
\multirow{2}{*}{CR} &
\multirow{2}{*}{TR} &
\multirow{2}{*}{DIMF} &
\multirow{2}{*}{ASFT} &
\multirow{2}{*}{PR} &
\multicolumn{3}{c|}{MSE$\downarrow$} &
$C_{\text{MSE}}$$\downarrow$ &
$CF_{\text{MSE}}$$\downarrow$ &
\multicolumn{3}{c|}{MSE$\downarrow$} &
$C_{\text{MSE}}$$\downarrow$ &
$CF_{\text{MSE}}$$\downarrow$ \\
\cmidrule(lr{0.4em}){6-8}
\cmidrule(l{0.8em}r{0.4em}){9-10}
\cmidrule(lr{0.4em}){11-13}
\cmidrule(l{0.8em}r{0.4em}){14-15}
& & & & &
0.5s & 1.0s & 1.5s & 1.5s & 1.5s &
0.5s & 1.0s & 1.5s & 1.5s & 1.5s \\
\midrule
$\times$ & \checkmark & \checkmark & \checkmark & \checkmark &
33 & 70 & 139 & 110 & 253 &
18 & 40 & 80 & 62 & 140 \\
\checkmark & $\times$ & \checkmark & \checkmark & \checkmark &
34 & 68 & 137 & 111 & 260 &
20 & 38 & 84 & 59 & 130 \\
\checkmark & \checkmark & $\times$ & \checkmark & \checkmark &
32 & 68 & 137 & 107 & 248 &
19 & 37 & 78  & 55 & 133 \\
\checkmark & \checkmark & \checkmark & $\times$ & \checkmark &
36 & 72 & 141 & 110 & 254 &
17 & 41 & 79 & 60 & 125 \\
\checkmark & \checkmark & \checkmark & \checkmark & $\times$ &
33 & 69 & 138 & 112 & 249 &
16 & 35 & 66  & 56 & 135 \\
\checkmark & \checkmark & \checkmark & \checkmark & \checkmark &
\textbf{29} & \textbf{65} & \textbf{130} & \textbf{101} & \textbf{242} &
\textbf{14} & \textbf{28} & \textbf{61}  & \textbf{40}  & \textbf{119} \\
\bottomrule
\end{tabular}
\end{table*}

\begin{table}[t]
\centering
\renewcommand{\arraystretch}{0.9}
\caption{Ablation Study of Key Components on JAAD.}
\label{Ablation_keycomp1}
\setlength{\tabcolsep}{3pt}
\fontsize{10}{10}\selectfont
\begin{tabular}{
>{\centering\arraybackslash}m{1.1cm}
>{\centering\arraybackslash}m{1.1cm}|
>{\hspace{4pt}}c<{\hspace{4pt}}
>{\hspace{4pt}}c<{\hspace{4pt}}
>{\hspace{4pt}}c<{\hspace{4pt}}|c c}
\toprule
\multirow{2}{*}{MMFE} &
\multirow{2}{*}{MEHP} &
\multicolumn{3}{c|}{MSE$\downarrow$} &
$C_{\text{MSE}}\downarrow$ & $CF_{\text{MSE}}\downarrow$ \\
\cmidrule(lr){3-5}\cmidrule(lr){6-7}
& & 0.5s & 1.0s & 1.5s & 1.5s & 1.5s \\
\midrule
$\checkmark$ & $\checkmark$ & \textbf{29} & \textbf{65} & \textbf{130} & \textbf{101} & \textbf{242} \\
$\checkmark$ & $\times$     & 38 & 80 & 152 & 121 & 265 \\
$\times$     & $\checkmark$ & 42 & 83 & 160 & 130 & 271 \\
\bottomrule
\end{tabular}
\end{table}

\begin{table}[t]
\centering
\renewcommand{\arraystretch}{0.9}
\caption{Ablation Study of Key Components on PIE.}
\label{Ablation_keycomp2}
\setlength{\tabcolsep}{3pt}
\fontsize{10}{10}\selectfont
\begin{tabular}{
>{\centering\arraybackslash}m{1.1cm}
>{\centering\arraybackslash}m{1.1cm}|
>{\hspace{4pt}}c<{\hspace{4pt}}
>{\hspace{4pt}}c<{\hspace{4pt}}
>{\hspace{4pt}}c<{\hspace{4pt}}|c c}
\toprule
\multirow{2}{*}{MMFE} &
\multirow{2}{*}{MEHP} &
\multicolumn{3}{c|}{MSE$\downarrow$} &
$C_{\text{MSE}}\downarrow$ & $CF_{\text{MSE}}\downarrow$ \\
\cmidrule(lr){3-5}\cmidrule(lr){6-7}
& & 0.5s & 1.0s & 1.5s & 1.5s & 1.5s \\
\midrule
$\checkmark$ & $\checkmark$ & \textbf{14} & \textbf{28} & \textbf{61} & \textbf{40} & \textbf{119} \\
$\checkmark$ & $\times$     & 22 & 48 & 81 & 62 & 145 \\
$\times$     & $\checkmark$ & 32 & 52 & 85 & 66 & 139 \\
\bottomrule
\end{tabular}
\end{table}

\subsubsection{Multiscale Feature Size}
\begin{table*}[!ht]
\centering
\renewcommand{\arraystretch}{1.0}
\setlength{\tabcolsep}{5pt}  
\fontsize{10}{10}\selectfont
\caption{Ablation study on the impact of scale size on the JAAD and PIE test sets. \textbf{Bold} is best. Lower is better.}
\label{ablation_scale}
\begin{tabular}{
  >{\centering\arraybackslash}p{0.8cm}
  >{\centering\arraybackslash}p{0.8cm}
  >{\centering\arraybackslash}p{0.8cm}
  >{\centering\arraybackslash}p{0.8cm}
  | >{\centering\arraybackslash}p{0.8cm}
    >{\centering\arraybackslash}p{0.8cm}
    >{\centering\arraybackslash}p{0.8cm}
  | >{\centering\arraybackslash}p{1.2cm}
    >{\centering\arraybackslash}p{1.2cm}
  | >{\centering\arraybackslash}p{0.8cm}
    >{\centering\arraybackslash}p{0.8cm}
    >{\centering\arraybackslash}p{0.8cm}
  | >{\centering\arraybackslash}p{1.2cm}
    >{\centering\arraybackslash}p{1.2cm}}
\toprule
\multicolumn{4}{c|}{Scale Size} &
\multicolumn{5}{c|}{JAAD} &
\multicolumn{5}{c}{PIE} \\
\cmidrule(lr){1-4}\cmidrule(lr){5-9}\cmidrule(lr){10-14}
\multirow{2}{*}{1} &
\multirow{2}{*}{3} &
\multirow{2}{*}{5} &
\multirow{2}{*}{15} &
\multicolumn{3}{c|}{MSE$\downarrow$} &
$C_{\text{MSE}}$$\downarrow$ &
$CF_{\text{MSE}}$$\downarrow$ &
\multicolumn{3}{c|}{MSE$\downarrow$} &
$C_{\text{MSE}}$$\downarrow$ &
$CF_{\text{MSE}}$$\downarrow$ \\
\cmidrule(lr{0.4em}){5-7}
\cmidrule(l{0.8em}r{0.4em}){8-9}
\cmidrule(lr{0.4em}){10-12}
\cmidrule(l{0.8em}r{0.4em}){13-14}
& & & &
0.5s & 1.0s & 1.5s & 1.5s & 1.5s &
0.5s & 1.0s & 1.5s & 1.5s & 1.5s \\
\midrule
\checkmark & $\times$ & \checkmark & \checkmark &
40 & 75 & 143 & 112 & 259 &
20 & 35 & 71  & 48 & 130 \\
\checkmark & \checkmark & $\times$ & \checkmark &
33 & 69 & 133 & 104 & 248 &
16 & 31 & 64 & 44 & 124 \\
\checkmark & \checkmark & \checkmark & $\times$ &
\textbf{29} & 68 & 133 & 102 & 244 &
16 & 29 & 63  & \textbf{40} & 121 \\
\checkmark & \checkmark & \checkmark & \checkmark &
\textbf{29} & \textbf{65} & \textbf{130} & \textbf{101} & \textbf{242} &
\textbf{14} & \textbf{28} & \textbf{61} & \textbf{40} & \textbf{119} \\
\bottomrule
\end{tabular}
\end{table*}

In the proposed MUSCLE-Net, input features are decomposed into four temporal scales $s_k \in \{1, 3, 5, 15\}$ to capture motion dynamics over different time horizons. This ablation study investigates the impact of scale selection by selectively removing individual temporal scales.

As reported in Table~\ref{ablation_scale}, the full multiscale configuration achieves the best or tied-best performance on both datasets. Removing fine-grained scales ($s=3$ or $s=5$) leads to performance degradation, indicating that high-resolution temporal cues are important for modeling rapid motion changes. Conversely, excluding the coarse scale ($s=15$) also harms performance, highlighting the importance of long-range temporal context in stabilizing predictions and reducing drift.

These results empirically demonstrate that a balanced multiscale hierarchy that integrates both local motion details and global temporal structure is essential for reliable trajectory and bounding-box prediction.

\subsubsection{Multiscale Feature Extracting Method}

\begin{figure}[t]\centering
    \hspace{-0.35cm}
    \includegraphics[width=8.5cm]{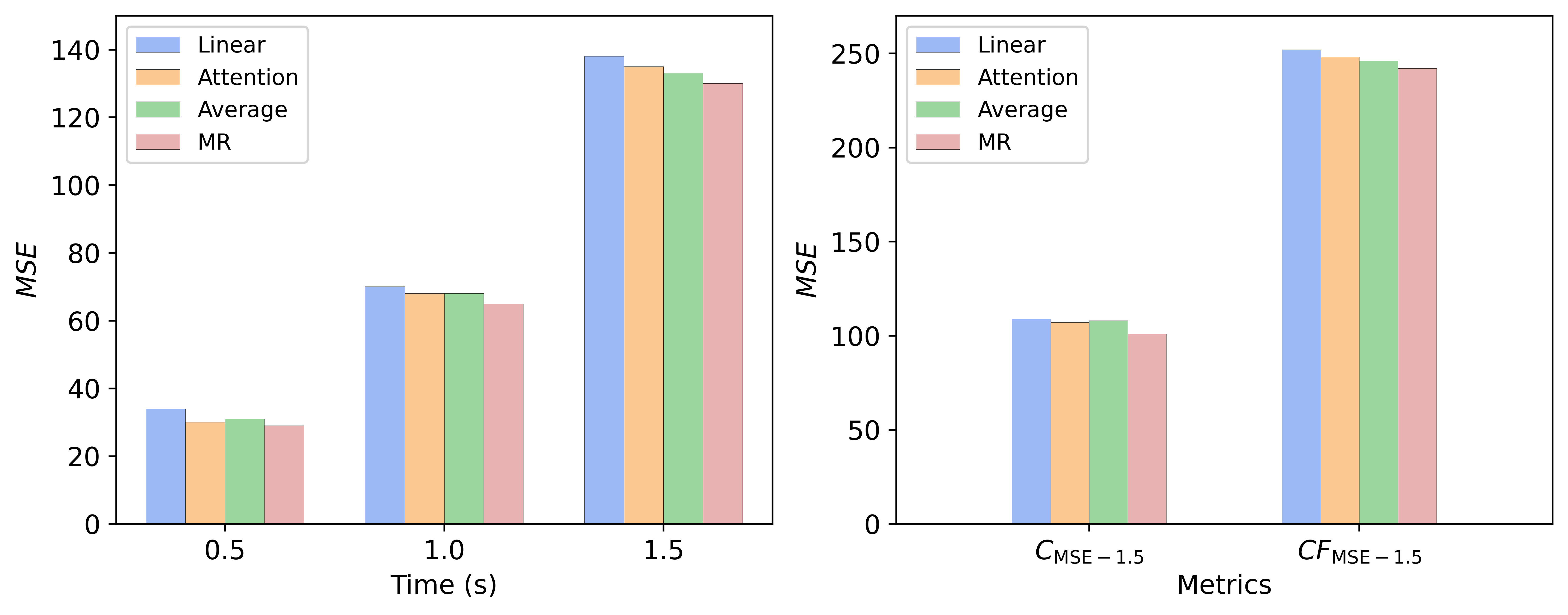}
    \includegraphics[width=8.5cm]{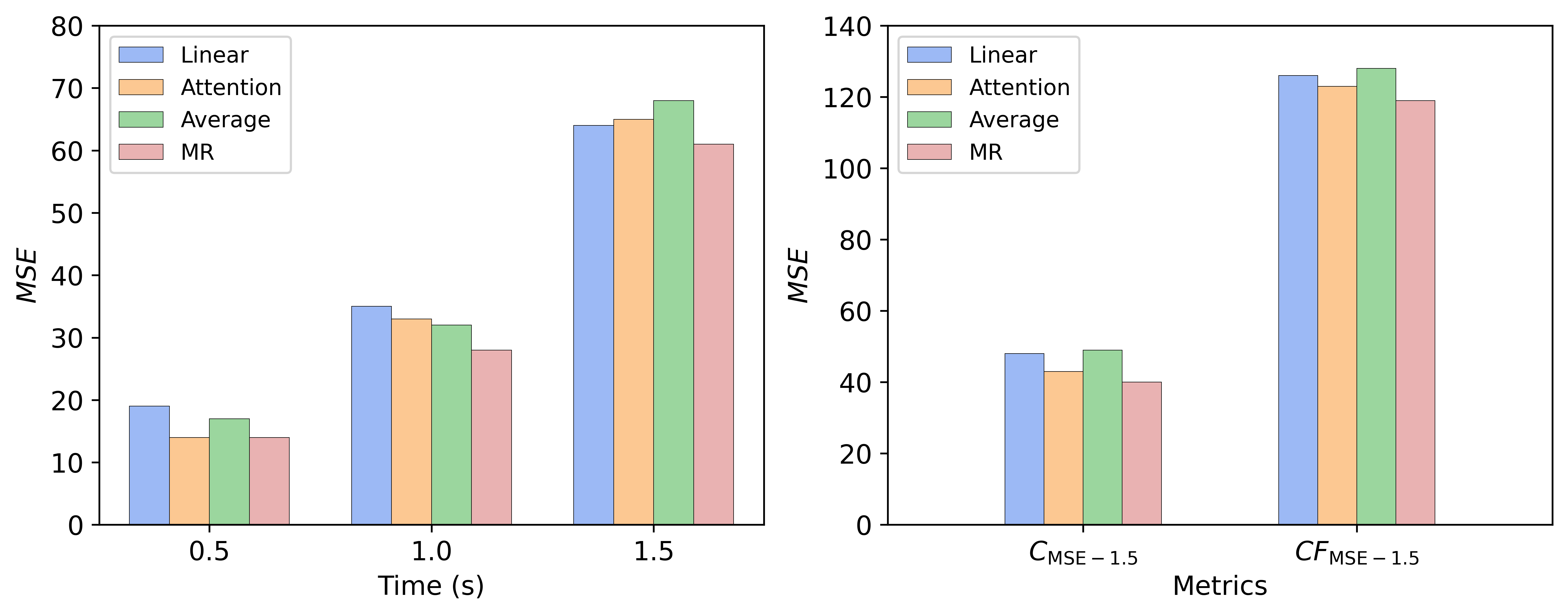}
	\caption{ Comparison of feature extraction methods applied to each temporal window. Results on the JAAD dataset are shown in the top row, while those on the PIE dataset are shown in the bottom row.}
    \label{Ablation_tpa}
\end{figure}

In MUSCLE-Net, we employ a Multiscale Encoder (ME) block to extract representations within each temporal window at multiple scales. To assess its effectiveness, we compare ME with several alternative aggregation strategies. The Linear variant compresses each window using a lightweight MLP without explicit attention. The Attention baseline applies a standard self-attention layer followed by an MLP to generate window-level features. The Average strategy represents each window by simple mean pooling.

As shown in Figure~\ref{Ablation_tpa}, ME consistently outperforms all baseline methods. Average pooling fails to capture fine-grained temporal variations, while the Linear variant lacks effective selective weighting across frames. Although standard attention improves performance to some extent, it is less stable and less effective at modeling multiscale motion compared with ME’s explicit window-wise key and value design. These results indicate that ME achieves a more favorable balance between representational expressiveness and training stability, thereby yielding more informative multiscale representations for downstream prediction tasks.

\subsubsection{Number of Coarse Sampling M}

\begin{figure}[t]\centering
    \hspace{-0.3cm}
	\includegraphics[width=8.3cm]{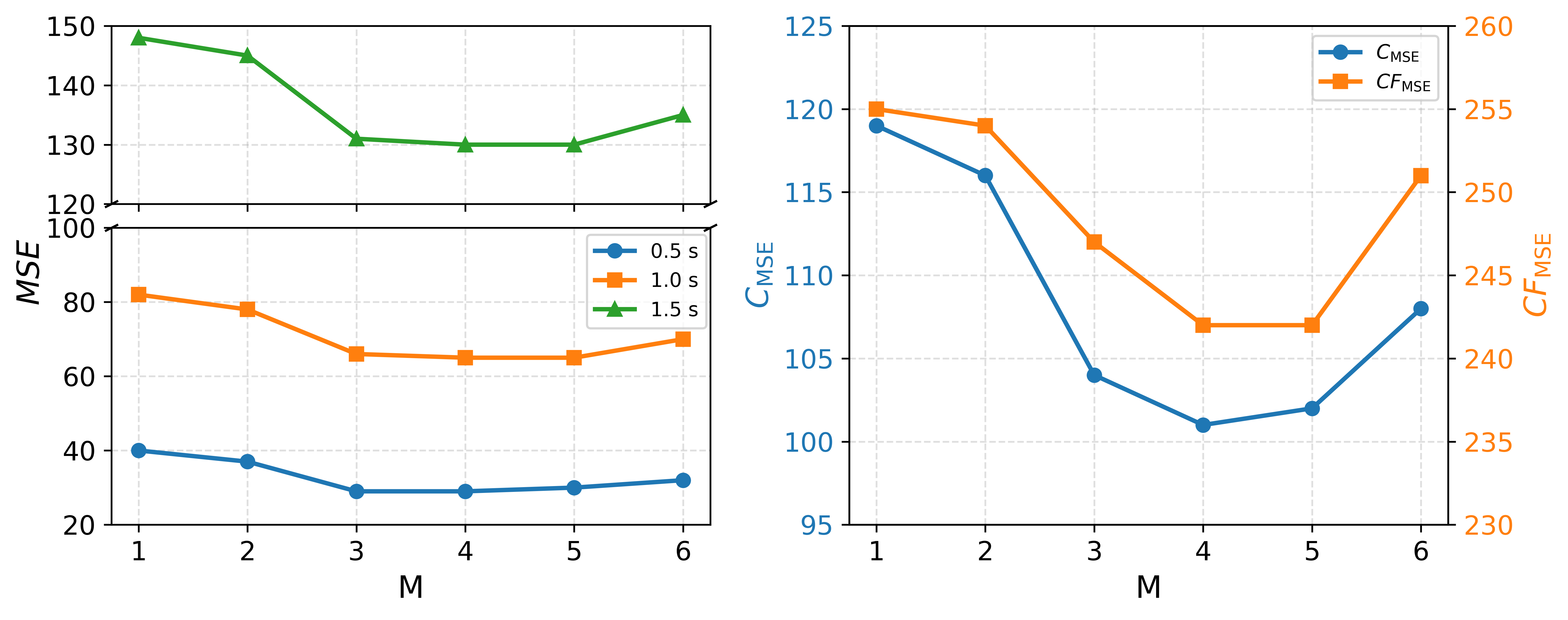}
    \includegraphics[width=8.3cm]{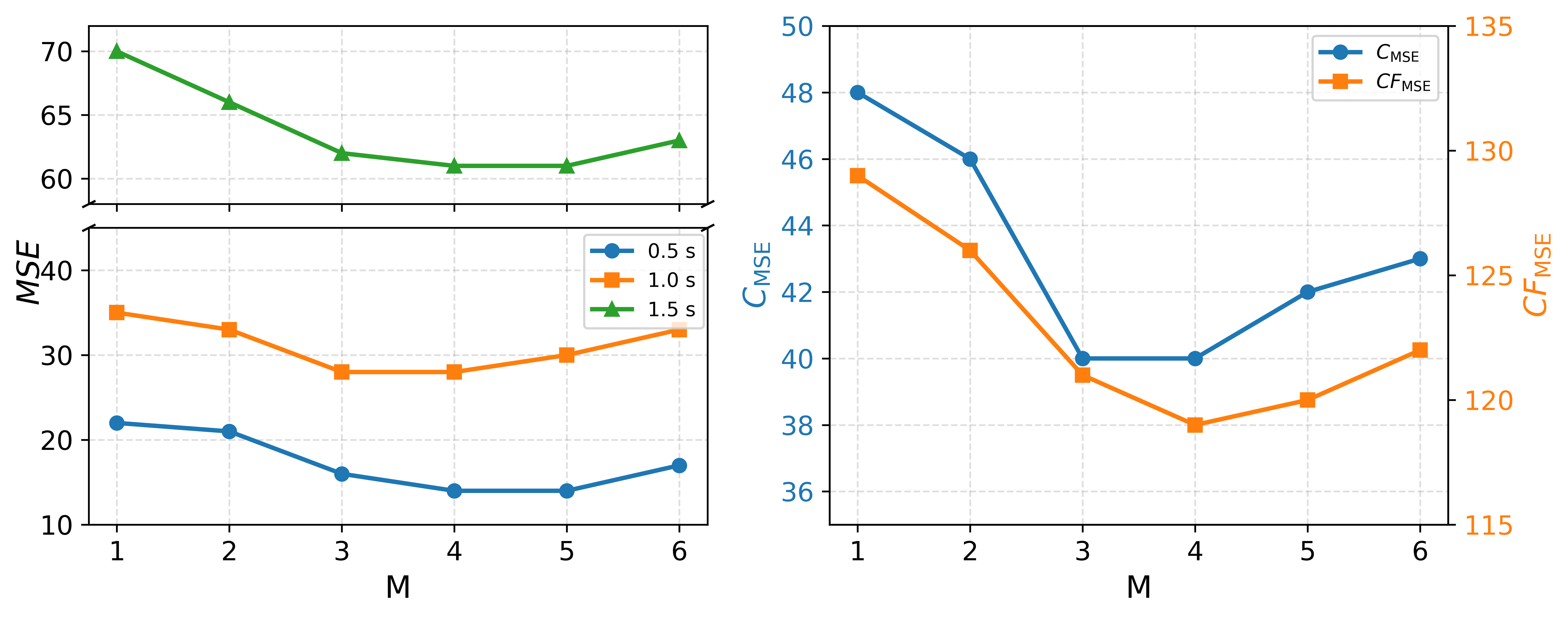}
	\caption{Comparison of sampling numbers M at the coarse stage. The top row shows results on the JAAD dataset, and the bottom row presents results on the PIE dataset. }
    \label{Ablation_m}
\end{figure}

We further analyze the effect of the sampling number $M$ in the coarse prediction stage by evaluating values from $M=1$ to $6$. As shown in Figure~\ref{Ablation_m}, performance improves consistently as $M$ increases from $1$ to $4$, suggesting that a larger set of coarse hypotheses enables better coverage of multimodal future trajectories. When $M$ exceeds $4$, no additional performance gains are observed, and a slight degradation occurs. This behavior is likely caused by increased variance in coarse predictions and reduced stability in the subsequent refinement stage. Based on this analysis, we select $M=4$ as a balanced and effective setting.

\subsubsection{Number of Refinement Anchor N}

\begin{figure}[t]\centering
    \hspace{-0.2cm}
	\includegraphics[width=8.5cm]{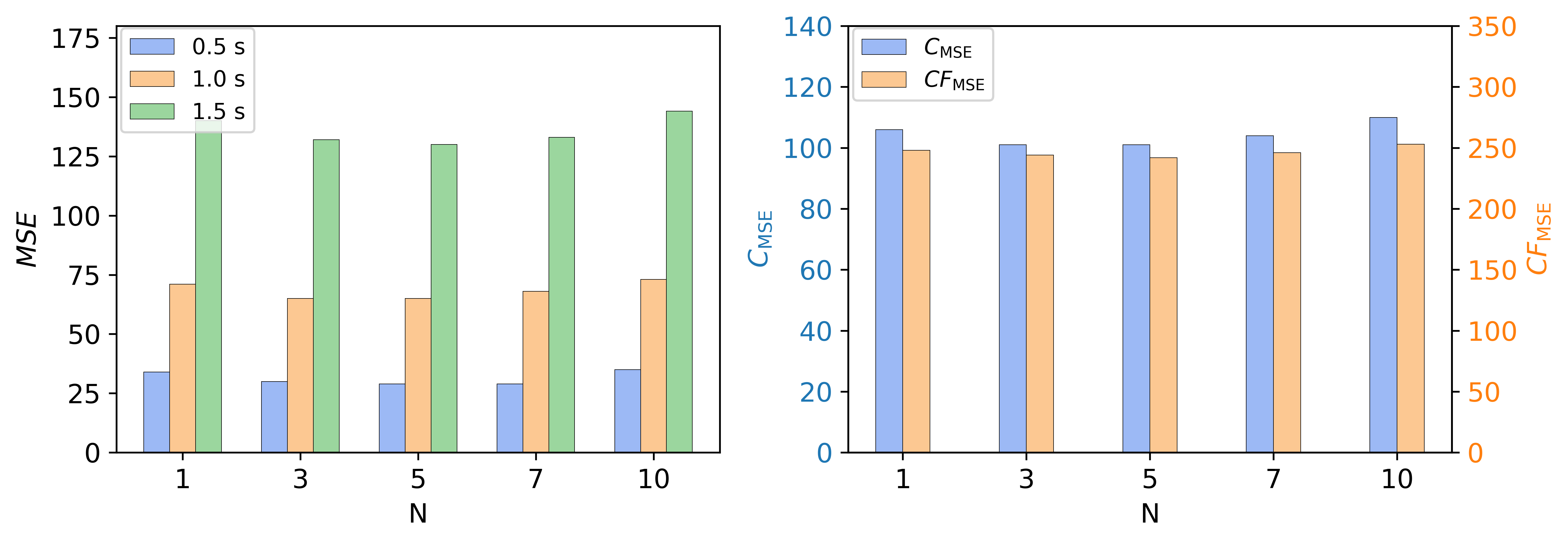}
    \includegraphics[width=8.5cm]{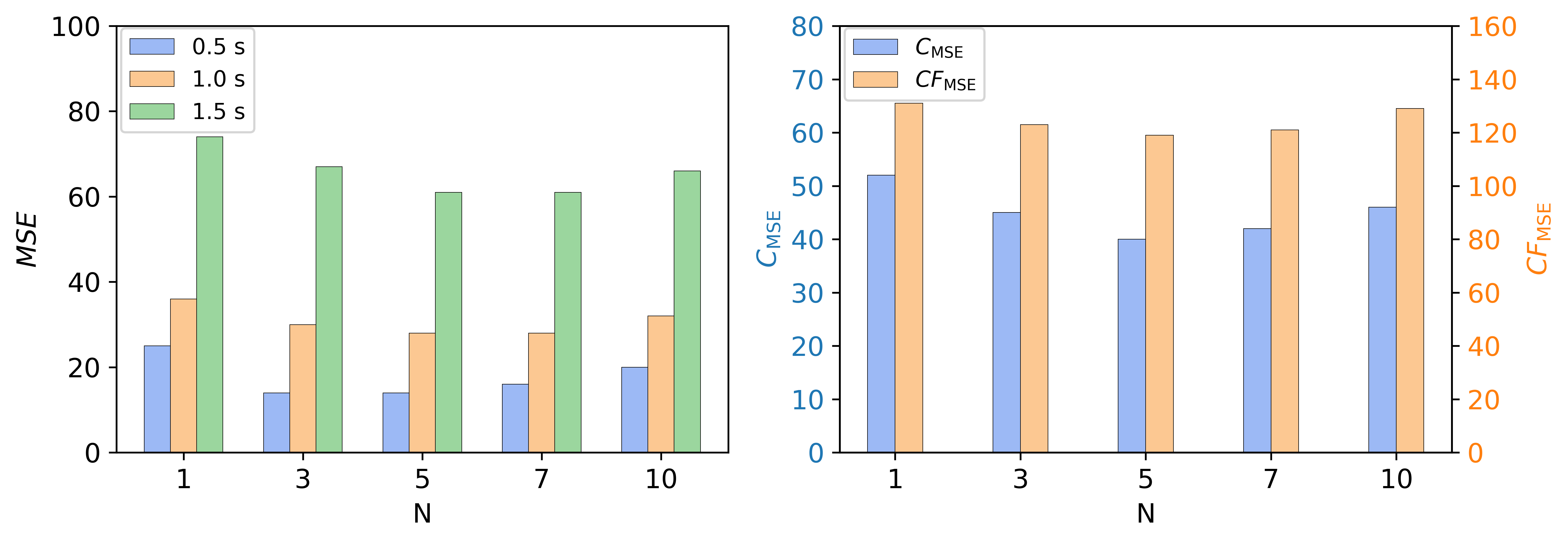}
	\caption{Comparison results for different numbers of refinement anchors $N$. The top row reports results on the JAAD dataset, while the bottom row shows results on the PIE dataset.}
    \label{Ablation_n}
\end{figure}

We evaluate the impact of the number of refinement anchors $N$ by testing $N \in \{1, 3, 5, 7, 10\}$. As illustrated in Figure~\ref{Ablation_n}, increasing $N$ improves performance up to $N=5$, beyond which no further gains are observed. Smaller values limit local correction diversity, while larger values introduce redundancy and destabilize refinement. Therefore, $N=5$ is adopted in our final model.

\subsubsection{Method of Coarse Prediction}
\begin{figure}[t]\centering
    \hspace{-0.3cm}
	\includegraphics[width=8.3cm]{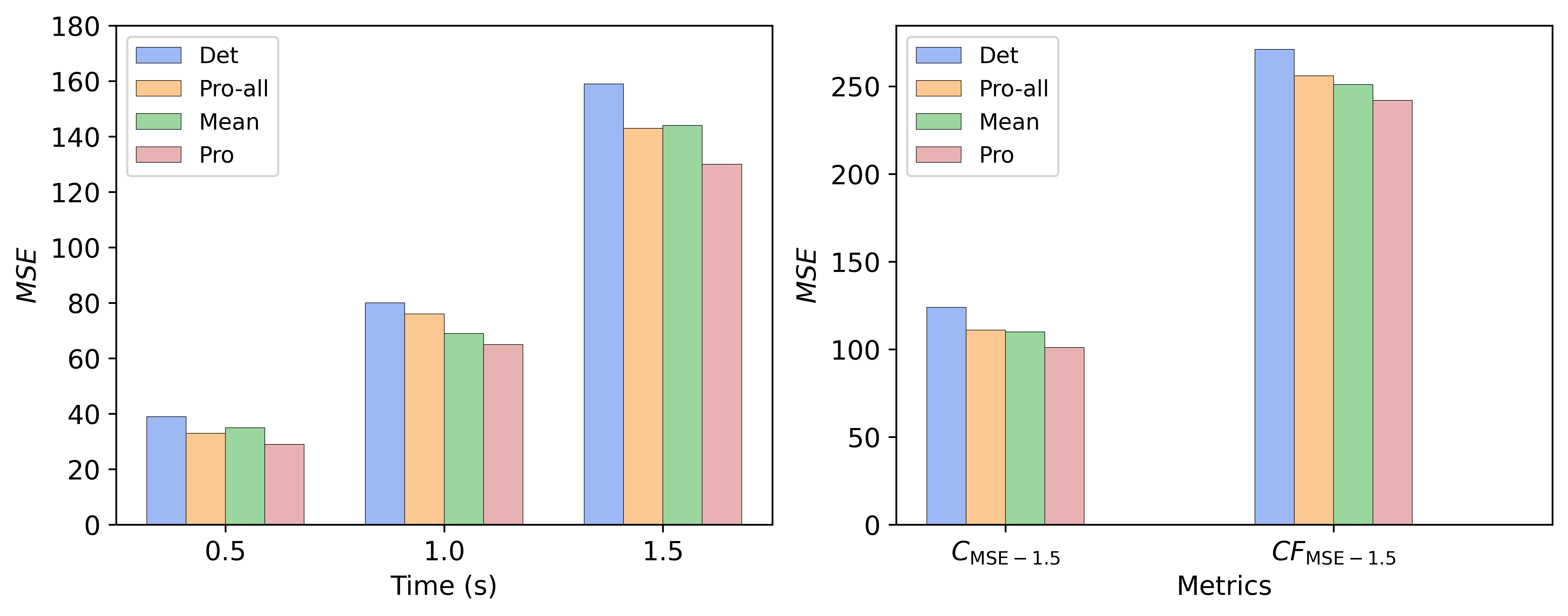}
    \hspace{0.2cm}
    \includegraphics[width=8.3cm]{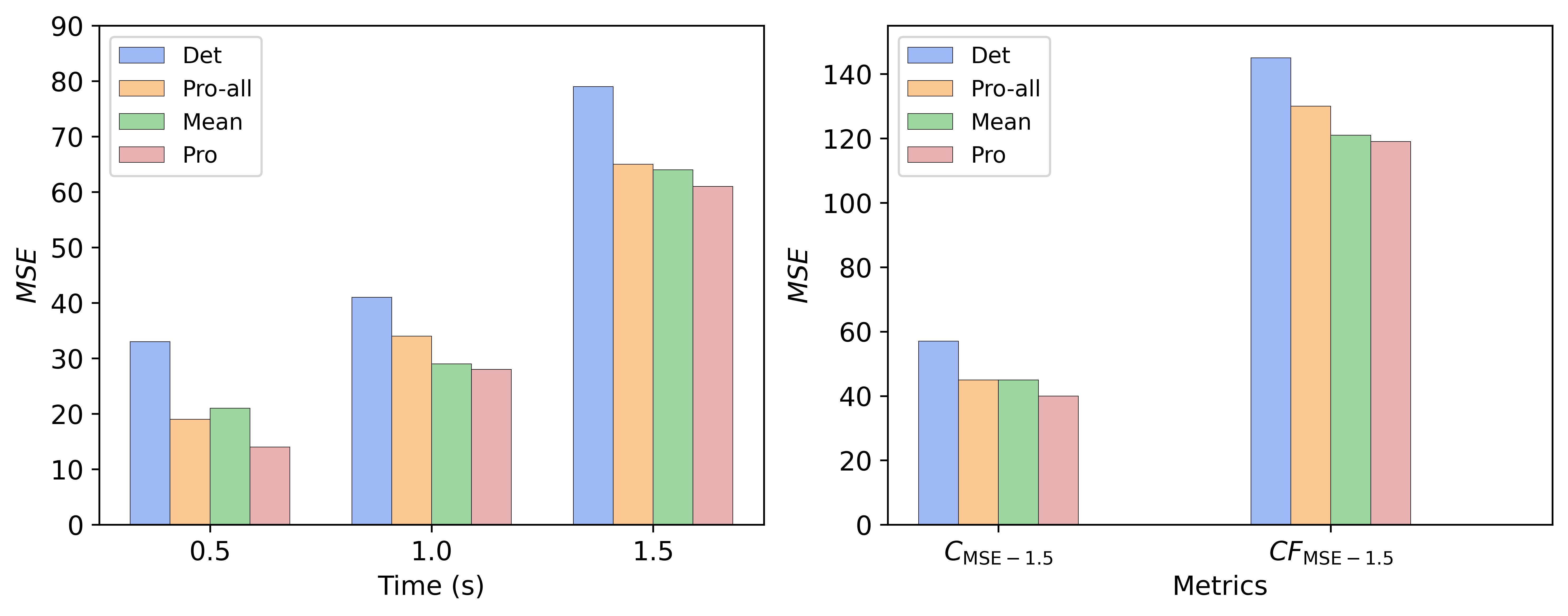}
	\caption{The comparison of different coarse prediction strategies is shown. Det denotes the deterministic variant; Pro-all optimizes over all sampled candidates instead of relying on a single one; and Mean uses the predicted mean without sampling. The top row presents results on the JAAD dataset, while the bottom row shows results on the PIE dataset.}
    \label{Ablation_coarse}
\end{figure}

We analyze the design of the coarse prediction module by comparing three variants to examine whether uncertainty-aware probabilistic initialization benefits subsequent hierarchical refinement. Deterministic: the coarse module predicts a single bounding-box trajectory and is supervised with an $L_2$ loss, without probabilistic modeling. Optimize-All: all sampled hypotheses are jointly optimized using the NLL objective, without selecting the best candidate for supervision. Mean only Gaussian: the model predicts the mean and variance of a Gaussian distribution and is trained with NLL, but no sampling is performed, and the predicted mean is used as the coarse output.

As shown in Figure~\ref{Ablation_coarse}, the deterministic variant yields the worst performance, indicating that uncertainty modeling is important for robust coarse prediction. Optimizing all hypotheses leads to overly smoothed results due to conflicting supervision signals. While the mean-only Gaussian variant improves stability, it still underperforms Gaussian-based coarse prediction with sampling. These results suggest that the Gaussian-based probabilistic coarse predictor is not introduced merely for distribution modeling, but to provide more informative uncertainty-aware coarse initialization for the subsequent refinement stage. In particular, sampling-induced diversity helps generate coarse hypotheses that better support refinement.

\subsubsection{ASFT Feature Align Method}
During the scale alignment stage, ASFT upsamples each coarse feature sequence to the original frame rate prior to fusion. Our default implementation employs linear interpolation, which preserves the temporal structure within each window. To assess the importance of this design choice, we evaluate two alternative alignment strategies. Nearest: each coarse feature is directly repeated across its corresponding window of $s_k$ frames, resulting in a piecewise constant sequence without modeling intra-window temporal variation. Window-average broadcast: each window feature is averaged and broadcast to all frames within the window, completely discarding fine-grained temporal dynamics.

\begin{figure}[t]\centering
    \hspace{-0.3cm}
	\includegraphics[width=8.5cm]{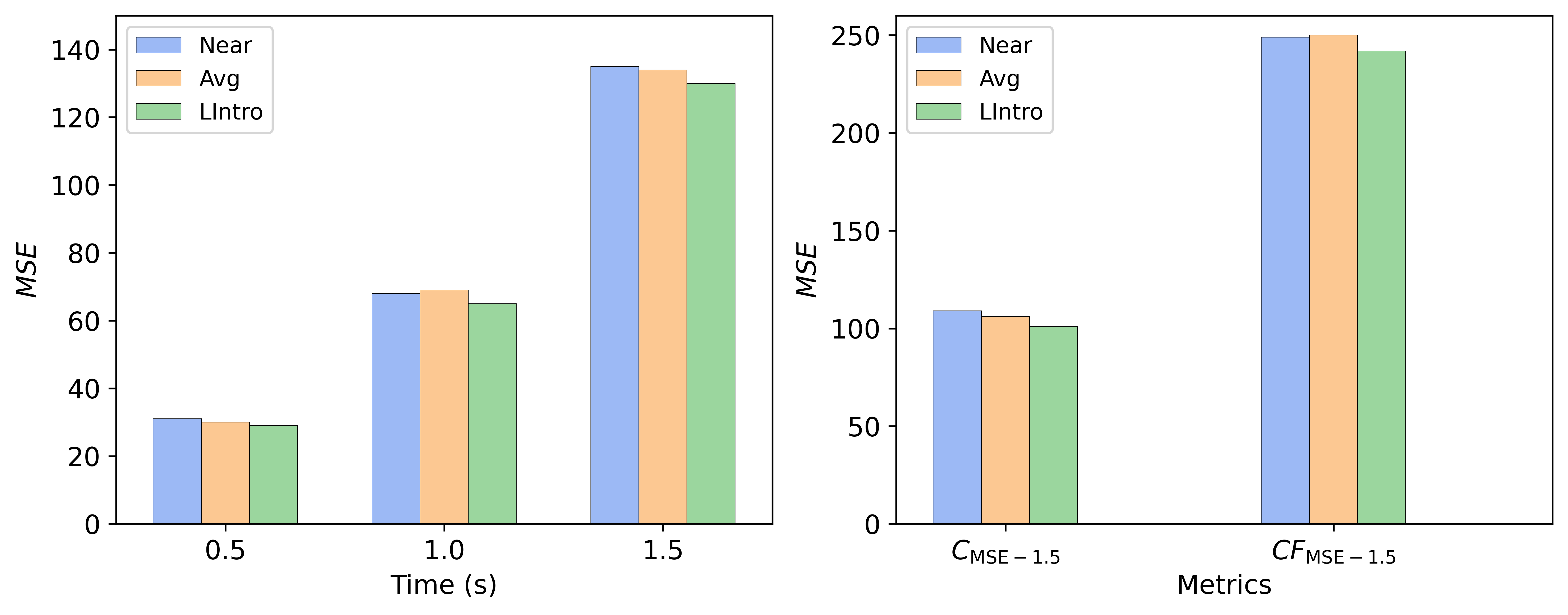}
    \hspace{0.2cm}
    \includegraphics[width=8.5cm]{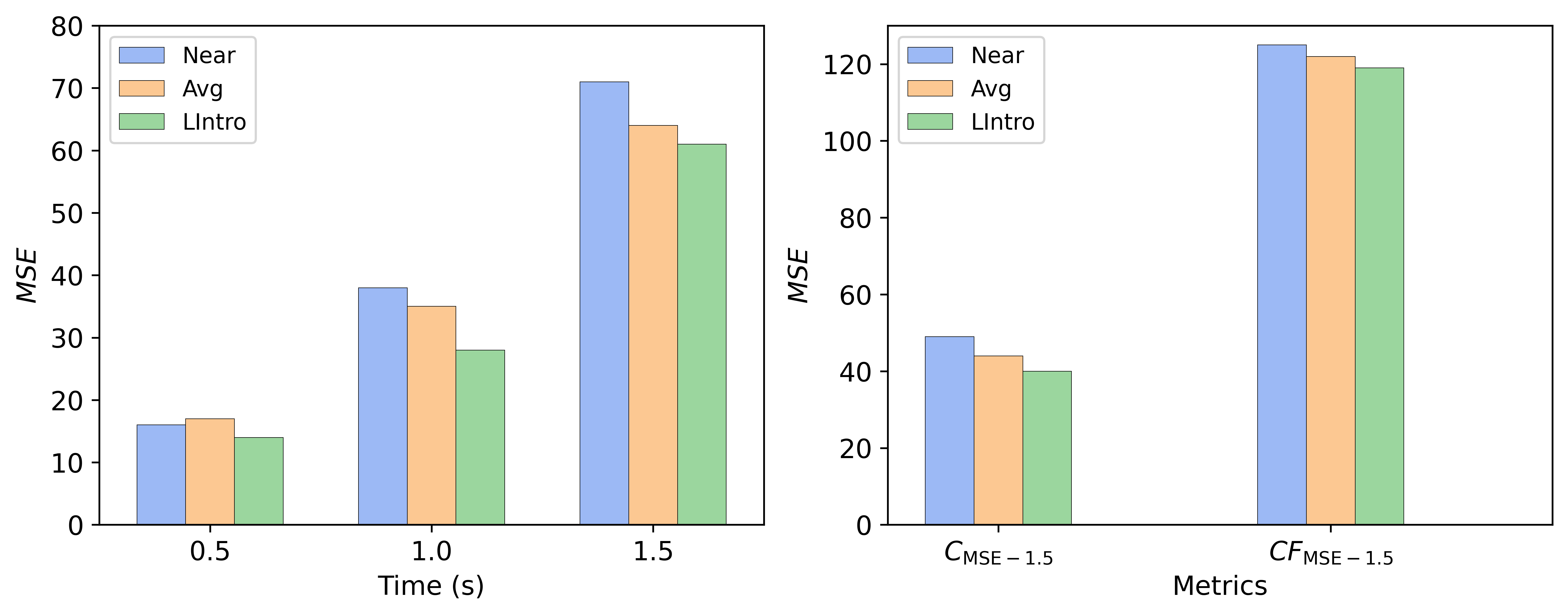}
	\caption{Comparison of different ASFT alignment strategies. Near repeats the value from the corresponding temporal window, Avg applies average pooling within the window, and LIntro performs linear interpolation. The top row presents results on the JAAD dataset, while the bottom row shows results on the PIE dataset.}
    \label{Ablation_asft_align}
\end{figure}

As shown in Figure~\ref{Ablation_asft_align}, both alternatives lead to performance degradation. The nearest strategy partially retains global motion trends but introduces abrupt temporal discontinuities, while window-average broadcast excessively smooths motion changes that are important for accurate prediction. In contrast, linear interpolation yields the best performance by maintaining smooth and structurally consistent temporal evolution during scale alignment.

\subsubsection{LCA Window Size k}
The PR module refines coarse predictions by aggregating information from a local temporal neighborhood. By default, we adopt a compact refinement window that considers one time step before and after each prediction. To analyze the effect of temporal context, we compare four settings: no local context, a one-step neighborhood, a two-step neighborhood, and a three-step neighborhood.

As shown in Tables~\ref{Ablation_lcak1} and~\ref{Ablation_lcak2}, removing local temporal context results in the poorest performance, highlighting the importance of short-term temporal structure for effective refinement. The one-step neighborhood achieves the best accuracy, offering sufficient local cues while avoiding unnecessary interference. Expanding the neighborhood further does not provide additional benefits and instead introduces less relevant context, which weakens the refinement’s focus on precise motion correction. These results indicate that a compact temporal window is most effective for progressive refinement.

\begin{table}[t]
\centering
\renewcommand{\arraystretch}{0.9}
\caption{Impact of LCA window size K on the JAAD Dataset.}
\label{Ablation_lcak1}
\setlength{\tabcolsep}{3pt}
\fontsize{10}{10}\selectfont
\begin{tabular}{
>{\centering\arraybackslash}m{1.8cm}| 
>{\hspace{4pt}}c<{\hspace{4pt}}
>{\hspace{4pt}}c<{\hspace{4pt}}
>{\hspace{4pt}}c<{\hspace{4pt}}|c c}
\toprule
\multirow{2}{*}{K} &
\multicolumn{3}{c|}{MSE$\downarrow$} &
$C_{\text{MSE}}$ $\downarrow$ & $CF_{\text{MSE}}$$\downarrow$ \\
\cmidrule(lr){2-4}\cmidrule(lr){5-6}
& 0.5s & 1.0s & 1.5s & 1.5s & 1.5s \\
\midrule
0  & 31 & 69 & 138  & 107 & 248 \\
1  & \textbf{29} & \textbf{65} & \textbf{130}  & \textbf{101} & \textbf{242} \\
2  & \textbf{29} & 67 & 131  & 103 & 245 \\
3 & 34 & 71 & 140  & 109 & 251 \\
\bottomrule
\end{tabular}
\end{table}

\begin{table}[t]
\centering
\renewcommand{\arraystretch}{0.9}
\caption{Impact of LCA window size K on the PIE Dataset.}
\label{Ablation_lcak2}
\setlength{\tabcolsep}{3pt}
\fontsize{10}{10}\selectfont
\begin{tabular}{
>{\centering\arraybackslash}m{1.8cm}| 
>{\hspace{4pt}}c<{\hspace{4pt}}
>{\hspace{4pt}}c<{\hspace{4pt}}
>{\hspace{4pt}}c<{\hspace{4pt}}|c c}
\toprule
\multirow{2}{*}{K} &
\multicolumn{3}{c|}{MSE$\downarrow$} &
$C_{\text{MSE}}$ $\downarrow$ & $CF_{\text{MSE}}$$\downarrow$ \\
\cmidrule(lr){2-4}\cmidrule(lr){5-6}
& 0.5s & 1.0s & 1.5s & 1.5s & 1.5s \\
\midrule
0  & \textbf{14} & 30 & 63  & 44 & 124 \\
1  & \textbf{14} & \textbf{28} & \textbf{61}  & \textbf{40} & \textbf{119} \\
2  & 17 & 33 & 65  & 46 & 126 \\
3  & 19 & 35 & 65  & 48 & 129 \\
\bottomrule
\end{tabular}
\end{table}

\subsubsection{Efficiency Analysis}

\begin{table}[h]
\centering
\caption{Module inference time under different batch sizes on JAAD.}
\setlength{\tabcolsep}{5pt}
\fontsize{10}{10}\selectfont
\begin{tabular}{ccccc}
\toprule
Batchsize & \makecell{MMFE \\ (ms)} & \makecell{MEHP \\ (ms)} & \makecell{Total \\ (ms/mb)} & \makecell{Params \\ (M)} \\
\midrule
32  & 8.21 & 5.46 & 13.67/26.36 & \multirow{4}{*}{0.151} \\
64  & 8.75 & 5.74 & 14.49/40.24 &                        \\
128 & 8.40 & 5.56 & 13.96/66.11  &                        \\
256 & 9.05 & 6.05 & 14.10/123.22 &                        \\
\bottomrule
\end{tabular}
\label{Ablation_speed}
\end{table}

\begin{table}[h]
\centering
\caption{Inference time and model size comparison on PIE.}
\fontsize{10}{10}\selectfont
\setlength{\tabcolsep}{8pt}
\begin{tabular*}{0.90\columnwidth}{@{\extracolsep{\fill}}lcc}
\toprule
Models                     & Time (ms) & Params (M) \\
\midrule
PIE$_{traj}$ \cite{ref71}  & 264.13 & 1.235 \\
MTN \cite{ref123}          & 7.32   & 0.134 \\
SGNet \cite{ref75}         & 321.14 & 7.621 \\
Ours                       & 11.12  & 0.151 \\
\bottomrule
\end{tabular*}
\label{Ablation_time}
\end{table}

To evaluate the computational cost of MUSCLE-Net, we report the module-level inference time under different batch sizes on JAAD and compare the inference time and model size with representative methods on PIE. As shown in Table~\ref{Ablation_speed}, the total inference time remains around 14 ms per batch across different batch sizes, while MMFE accounts for a larger portion of the runtime than MEHP. This indicates that the hierarchical prediction and refinement process introduces only moderate additional cost. As shown in Table~\ref{Ablation_time}, MUSCLE-Net has 0.151M parameters and achieves 11.12 ms inference time on PIE. Compared with PIE$_{traj}$ and SGNet, the proposed method is more efficient in both inference time and model size, while remaining close to the lightweight MTN. These results suggest that MUSCLE-Net improves prediction performance with a compact model size and moderate inference overhead.

\subsubsection{Missing and Degraded Modalities}

To evaluate robustness under incomplete or degraded observations, we conduct two additional experiments on PIE. We first remove one modality to simulate missing inputs, and then inject Gaussian noise into each modality to test degraded inputs.

As shown in Table~\ref{Ablation_missingModa}, removing velocity or pose information leads to higher errors than the full model, indicating that these modalities can provide complementary cues for trajectory prediction. The removal of velocity results in a relatively larger degradation at the 1.0s and 1.5s horizons, which suggests that motion dynamics may help preserve long-term trajectory consistency. Removing pose information also increases the prediction error, implying that pose cues may contribute to local motion modeling. Table \ref{Ablation_noiseModa} shows that prediction errors generally increase as the Gaussian noise level grows. Bounding-box perturbation tends to cause larger degradation under medium and large noise levels, while velocity and pose perturbations also affect performance in some horizons. These results suggest that the model shows gradual performance changes under modality-specific perturbations.

\begin{table}[!ht]
\centering
\renewcommand{\arraystretch}{0.9}
\caption{Impact of missing modality on the PIE Dataset.}
\label{Ablation_missingModa}
\setlength{\tabcolsep}{3pt}
\fontsize{10}{10}\selectfont
\begin{tabular}{
>{\centering\arraybackslash}m{1.8cm}| 
>{\hspace{4pt}}c<{\hspace{4pt}}
>{\hspace{4pt}}c<{\hspace{4pt}}
>{\hspace{4pt}}c<{\hspace{4pt}}|c c}
\toprule
\multirow{2}{*}{Method} &
\multicolumn{3}{c|}{MSE$\downarrow$} &
$C_{\text{MSE}}$ $\downarrow$ & $CF_{\text{MSE}}$$\downarrow$ \\
\cmidrule(lr){2-4}\cmidrule(lr){5-6}
& 0.5s & 1.0s & 1.5s & 1.5s & 1.5s \\
\midrule
Full  & 14 & 28 & 61  & 40 & 119 \\
w/o V & 16 & 35 & 73  & 52 & 140 \\
w/o P & 17 & 32 & 70  & 48 & 134 \\
\bottomrule
\end{tabular}
\end{table}


\begin{table}[!ht]
\centering
\renewcommand{\arraystretch}{0.9}
\caption{Robustness to modality-specific Gaussian noise on the PIE dataset. Each entry reports the result under bounding-box/velocity/pose perturbation.}
\label{Ablation_noiseModa}
\setlength{\tabcolsep}{3pt}
\fontsize{10}{10}\selectfont
\begin{tabular}{
>{\centering\arraybackslash}m{2.0cm}| 
>{\hspace{6pt}}c<{\hspace{6pt}}
>{\hspace{6pt}}c<{\hspace{6pt}}
>{\hspace{6pt}}c<{\hspace{6pt}}}
\toprule
\multirow{2}{*}{Gaussian std} &
\multicolumn{3}{c}{MSE$\downarrow$} \\
\cmidrule(lr){2-4}
& 0.5s & 1.0s & 1.5s \\
\midrule
0   & 14 & 28 & 61 \\
\midrule
2.5  & 15/14/14 & 31/29/34 & 65/63/68 \\
5    & 17/17/16 & 36/33/38 & 75/69/71 \\
10   & 22/19/24 & 45/39/43 & 81/78/76 \\
\bottomrule
\end{tabular}
\end{table}

\subsection{Scale Analysis}
\begin{figure}[!ht]\centering
	\includegraphics[width=8.8cm]{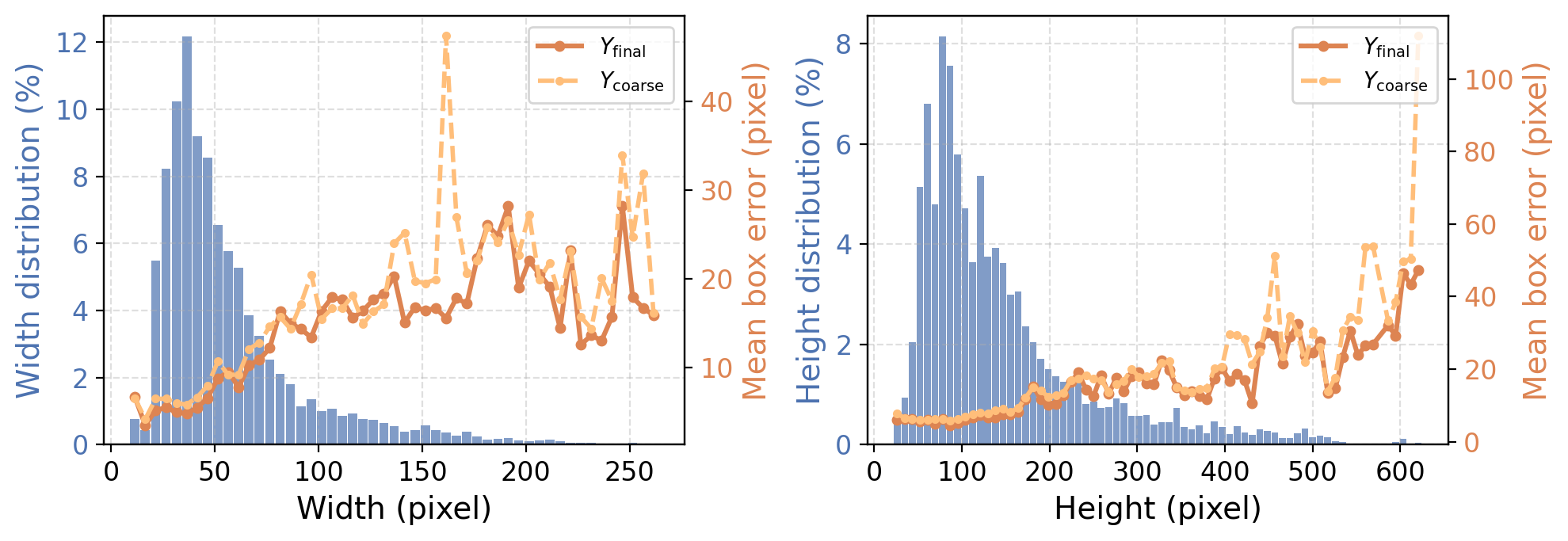}
    \includegraphics[width=8.8cm]{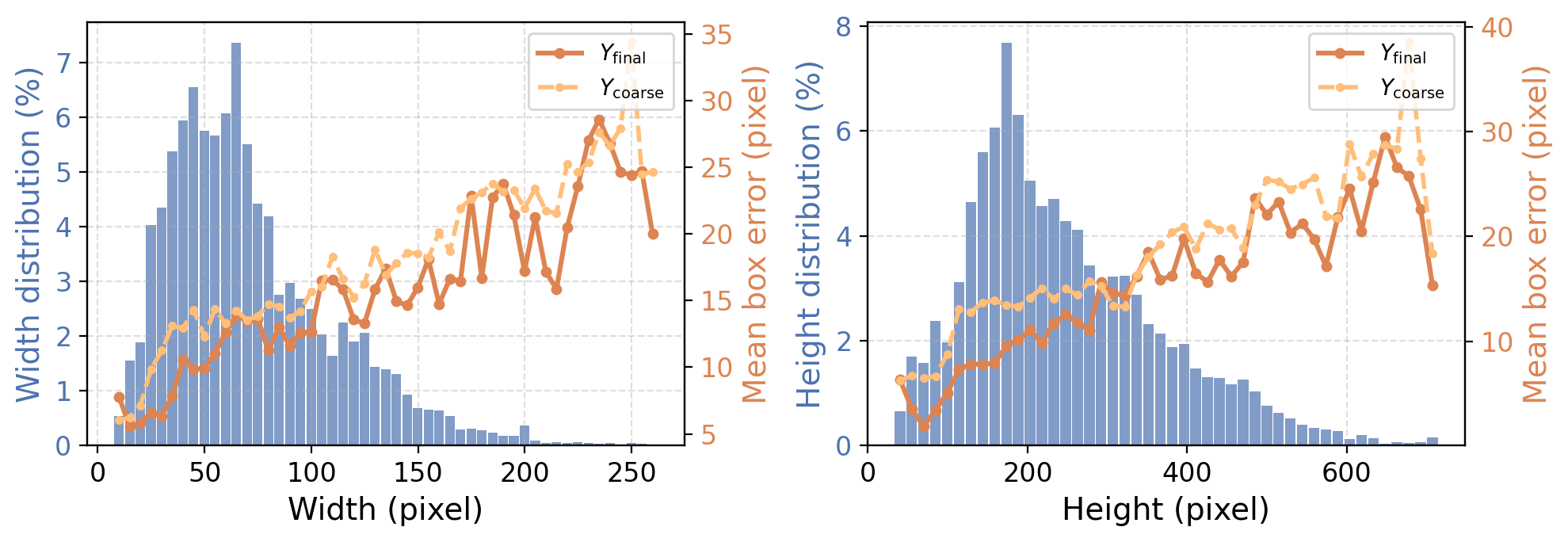}
	\caption{
   The distribution of bounding-box scale versus prediction error. The left and right columns correspond to width and height, respectively, with JAAD results shown in the top row and PIE results in the bottom row. }
    \label{vis_scale}
\end{figure}

Figure~\ref{vis_scale} illustrates the relationship between bounding-box scale and prediction error. On JAAD, most bounding-box widths lie in the range of $[20, 100]$ pixels and heights in $[30, 200]$ pixels, which is consistent with the distribution observed on PIE. Prediction errors tend to increase with larger bounding-box scales. One reason is that at close range, pixel-level sensitivity amplifies small localization deviations, leading to higher errors for larger boxes. In addition, the relatively limited number of training samples at large scales further constrains the model’s generalization. Notably, across all scales, the errors of $Y_{\text{final}}$ are consistently lower than those of $Y_{\text{coarse}}$, providing additional evidence of the effectiveness of the refinement stage.

\subsection{Failure Cases}
\begin{figure}[t]\centering
	\includegraphics[width=8.8cm]{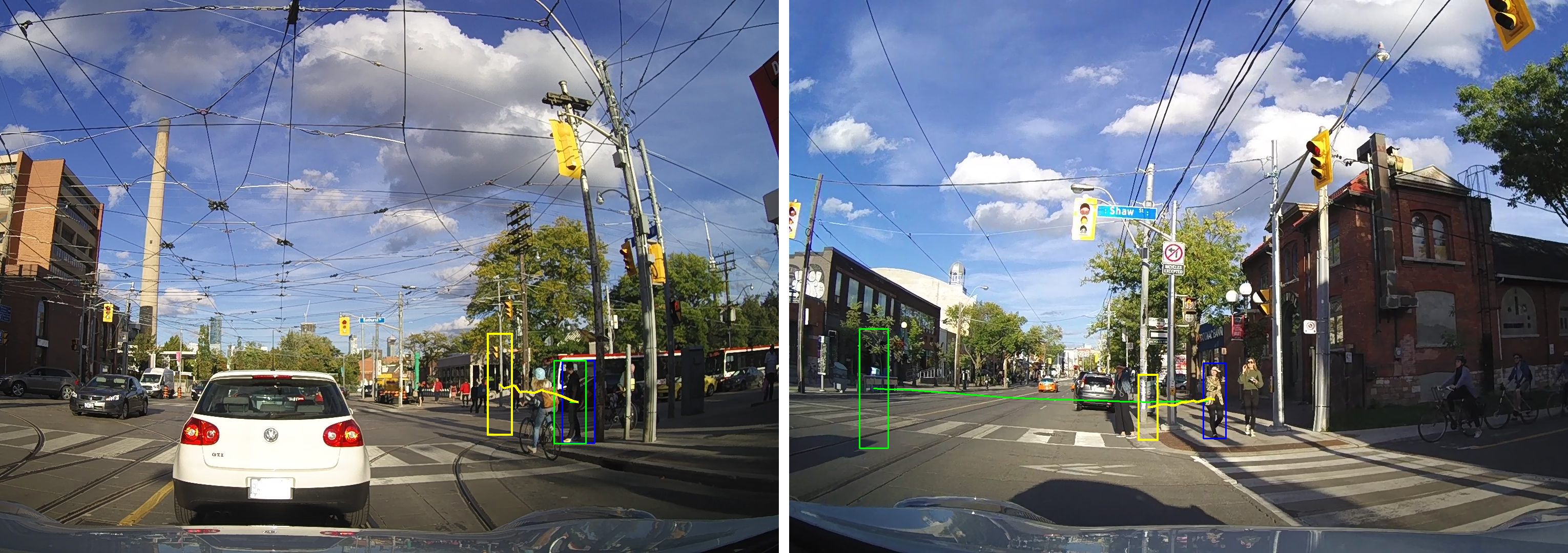}
	\caption{Representative failure cases of MUSCLE-Net.} 
    \label{Failure}
\end{figure}

Figure~\ref{Failure} presents two representative failure cases. In the left example, the pedestrian suddenly stops near the curb, deviating from the observed motion pattern and leading to an inaccurate prediction. Since the model primarily relies on historical motion cues, such abrupt behavioral changes are difficult to anticipate. In the right example, the ego vehicle begins to turn at the intersection, which alters the scene geometry and the relative positions of pedestrians. As the current model does not explicitly account for ego-motion direction or dynamic scene context, the predicted trajectory exhibits a larger deviation from the ground truth. These examples suggest that abrupt behavioral changes and dynamic scene variations remain challenging for trajectory prediction models. Future work could incorporate richer contextual cues and explicit ego-motion modeling to improve robustness in such scenarios.

\section{Conclusion}

In this work, we propose MUSCLE-Net, a Predicted-Multiscale-Aware Network for Pedestrian Trajectory Forecasting that exploits multimodal observations and scale-adaptive prediction. The framework consists of an MMFE module for calibrated multimodal representation learning and an MEHP module for coarse-to-fine trajectory estimation through multiscale fusion and progressive refinement. Experiments on benchmark datasets demonstrate the effectiveness and robustness of the proposed method, with consistent improvements across different motion patterns and scene settings. Although the current evaluation focuses on FPV traffic datasets aligned with the target ITS scenario, extending the proposed framework to top-down benchmarks for broader cross-view validation remains an important direction for future study. In future work, we plan to incorporate richer semantic and social context cues, such as scene layouts and interaction representations, and to explore stronger generative prediction strategies for better modeling highly multimodal futures in complex real-world environments.

\bibliographystyle{ieeetr}

\bibliography{refs}

@INPROCEEDINGS{ref1,
  author={Li, Kunming and Shan, Mao and Narula, Karan and Worrall, Stewart and Nebot, Eduardo},
  booktitle={Proceedings of the IEEE International Conference on Intelligent Transportation Systems}, 
  title={Socially Aware Crowd Navigation with Multimodal Pedestrian Trajectory Prediction for Autonomous Vehicles}, 
  year={2020},
  pages={1-8}
}

@INPROCEEDINGS{ref2,
  author={Liu, Yu and Zhang, Yuexin and Li, Kunming and Qiao, Yongliang and Worrall, Stewart and Li, You-Fu and Kong, He},
  booktitle={Proceedings of the IEEE International Conference on Intelligent Transportation Systems }, 
  title={Knowledge-aware Graph Transformer for Pedestrian Trajectory Prediction}, 
  year={2023},
  volume={},
  number={},
  pages={4360-4366},
  doi={10.1109/ITSC57777.2023.10421989}}

@INPROCEEDINGS{ref4,
  author={Eiffert, Stuart and Kong, He and Pirmarzdashti, Navid and Sukkarieh, Salah},
  booktitle={Proceedings of the IEEE International Conference on Robotics and Automation }, 
  title={Path Planning in Dynamic Environments using Generative RNNs and Monte Carlo Tree Search}, 
  year={2020},
  pages={10263-10269}
}

@article{ref11,
  title={Gaussian processes for machine learning},
  author={Seeger, Matthias},
  journal={{I}nternational {J}ournal of {N}eural {S}ystems},
  volume={14},
  number={02},
  pages={69--106},
  year={2004},
  publisher={World Scientific}
}

@INPROCEEDINGS{ref12,
  author={Lefèvre, Stéphanie and Laugier, Christian and Ibañez-Guzmán, Javier},
  booktitle={Proceedings of the IEEE Intelligent Vehicles Symposium (IV)}, 
  title={Exploiting map information for driver intention estimation at road intersections}, 
  year={2011},
  volume={},
  number={},
  pages={583-588},
  doi={10.1109/IVS.2011.5940452}}

@ARTICLE{ref13,
  author={Lefkopoulos, Vasileios and Menner, Marcel and Domahidi, Alexander and Zeilinger, Melanie N.},
  journal={IEEE Robotics and Automation Letters}, 
  title={Interaction-Aware Motion Prediction for Autonomous Driving: A Multiple Model Kalman Filtering Scheme}, 
  year={2021},
  volume={6},
  number={1},
  pages={80-87},
  doi={10.1109/LRA.2020.3032079}}

@INPROCEEDINGS{ref14,
  author={Alahi, Alexandre and Goel, Kratarth and Ramanathan, Vignesh and Robicquet, Alexandre and Fei-Fei, Li and Savarese, Silvio},
  booktitle={Proceedings of the IEEE/CVF Conference on Computer Vision and Pattern Recognition}, 
  title={Social {LSTM}: {H}uman Trajectory Prediction in Crowded Spaces}, 
  year={2016},
  volume={},
  number={},
  pages={961-971},
  doi={10.1109/CVPR.2016.110}}

@INPROCEEDINGS{ref15,
  author={Gupta, Agrim and Johnson, Justin and Fei-Fei, Li and Savarese, Silvio and Alahi, Alexandre},
  booktitle={Proceedings of the IEEE/CVF Conference on Computer Vision and Pattern Recognition}, 
  title={Social {GAN}: Socially Acceptable Trajectories with Generative Adversarial Networks}, 
  year={2018},
  volume={},
  number={},
  pages={2255-2264},
  doi={10.1109/CVPR.2018.00240}}

@ARTICLE{ref16,
  author={Eiffert, Stuart and Li, Kunming and Shan, Mao and Worrall, Stewart and Sukkarieh, Salah and Nebot, Eduardo},
  journal={IEEE Robotics and Automation Letters}, 
  title={Probabilistic Crowd {GAN}: Multimodal Pedestrian Trajectory Prediction Using a Graph Vehicle-Pedestrian Attention Network}, 
  year={2020},
  volume={5},
  number={4},
  pages={5026-5033},
  doi={10.1109/LRA.2020.3004324}}

@INPROCEEDINGS{ref27,
  author={Sadeghian, Amir and Kosaraju, Vineet and Sadeghian, Ali and Hirose, Noriaki and Rezatofighi, Hamid and Savarese, Silvio},
  booktitle={Proceedings of the IEEE/CVF Conference on Computer Vision and Pattern Recognition}, 
  title={SoPhie: An Attentive {GAN} for Predicting Paths Compliant to Social and Physical Constraints}, 
  year={2019},
  volume={},
  number={},
  pages={1349-1358},
  doi={10.1109/CVPR.2019.00144}}

@article{ref60, 
title={A Set of Control Points Conditioned Pedestrian Trajectory Prediction}, 
volume={37}, 
number={5}, 
journal={Proceedings of the AAAI Conference on Artificial Intelligence}, author={Bae, Inhwan and Jeon, Hae-Gon}, 
year={2023}, 
month={Jun.}, 
pages={6155-6165} 
}

@ARTICLE{ref61,
  author={Xie, Jiajia and Zhang, Sheng and Xia, Beihao and Xiao, Zhu and Jiang, Hongbo and Zhou, Siwang and Qin, Zheng and Chen, Hongyang},
  journal={IEEE Transactions on Multimedia}, 
  title={Pedestrian Trajectory Prediction Based on Social Interactions Learning With Random Weights}, 
  year={2024},
  volume={26},
  number={},
  pages={7503-7515}
}

@inproceedings{ref63,
  author={Shi, Liushuai and Wang, Le and Zhou, Sanping and Hua, Gang},
  booktitle={Proceedings of the IEEE/CVF International Conference on Computer Vision}, 
  title={Trajectory Unified Transformer for Pedestrian Trajectory Prediction}, 
  year={2023},
  pages={9641-9650}
}

@ARTICLE{ref64,
  author={Zhou, Xiangzheng and Chen, Xiaobo and Yang, Jian},
  journal={IEEE Transactions on Intelligent Transportation Systems}, 
  title={Edge-Enhanced Heterogeneous Graph Transformer With Priority-Based Feature Aggregation for Multi-Agent Trajectory Prediction}, 
  year={2025},
  volume={26},
  number={2},
  pages={2266-2281}
}

@ARTICLE{ref65,
  author={Chen, Xiaobo and Zhang, Huanjia and Deng, Fuwen and Liang, Jun and Yang, Jian},
  journal={IEEE Transactions on Intelligent Transportation Systems}, 
  title={Stochastic Non-Autoregressive Transformer-Based Multi-Modal Pedestrian Trajectory Prediction for Intelligent Vehicles}, 
  year={2024},
  volume={25},
  number={5},
  pages={3561-3574}
}

@ARTICLE{ref66,
  author={Hu, Chuan and Niu, Ruochen and Lin, Yiwei and Yang, Biao and Chen, Hao and Zhao, Baixuan and Zhang, Xi},
  journal={IEEE Transactions on Intelligent Transportation Systems}, 
  title={Probabilistic Trajectory Prediction of Vulnerable Road User Using Multimodal Inputs}, 
  year={2025},
  volume={26},
  number={2},
  pages={2679-2689},
  doi={10.1109/TITS.2024.3503683}}

@inproceedings{ref67,
      title={Stochastic Trajectory Prediction via Motion Indeterminacy Diffusion},
      author={Gu, Tianpei and Chen, Guangyi and Li, Junlong and Lin, Chunze and Rao, Yongming and Zhou, Jie and Lu, Jiwen},
      booktitle={Proceedings of the IEEE/CVF Conference on Computer Vision and Pattern Recognition},
      pages={17113--17122},
      year={2022}
    }

@inproceedings{ref68,
    author={Rempe, Davis and Luo, Zhengyi and Peng, Xue Bin and Yuan, Ye and Kitani, Kris and Kreis, Karsten and Fidler, Sanja and Litany, Or},
    title={Trace and Pace: Controllable Pedestrian Animation via Guided Trajectory Diffusion},
    booktitle={Proceedings of the IEEE/CVF Conference on Computer Vision and Pattern Recognition},
    pages={13756--13766},
    year={2023}
}

@INPROCEEDINGS{ref71,
  author={Rasouli, Amir and Kotseruba, Iuliia and Kunic, Toni and Tsotsos, John},
  booktitle={Proceedings of the IEEE/CVF International Conference on Computer Vision}, 
  title={PIE: A Large-Scale Dataset and Models for Pedestrian Intention Estimation and Trajectory Prediction}, 
  year={Oct. 2019},
  pages={6261-6270}
}

@INPROCEEDINGS{ref72,
  author={Rasouli, Amir and Kotseruba, Iuliia and Tsotsos, John K.},
  booktitle={Proceedings of the IEEE/CVF International Conference on Computer Vision Workshops (ICCVW)}, 
  title={Are They Going to Cross? A Benchmark Dataset and Baseline for Pedestrian Crosswalk Behavior}, 
  year={Oct. 2017},
  pages={206-213}
}

@ARTICLE{ref73,
  author={Dong, Yonghao and Wang, Le and Zhou, Sanping and Hua, Gang and Sun, Changyin},
  journal={IEEE Transactions on Multimedia}, 
  title={Sparse Pedestrian Character Learning for Trajectory Prediction}, 
  year={2024},
  volume={26},
  pages={11070-11082}
}

@inproceedings{ref74,
  title={A novel benchmarking paradigm and a scale-and motion-aware model for egocentric pedestrian trajectory prediction},
  author={Rasouli, Amir},
  booktitle={Proceedings of the IEEE International Conference on Robotics and Automation},
  pages={5630--5636},
  year={2024}
}

@article{ref75,
  title={Stepwise goal-driven networks for trajectory prediction},
  author={Wang, Chuhua and Wang, Yuchen and Xu, Mingze and Crandall, David J},
  journal={IEEE Robotics and Automation Letters},
  volume={7},
  number={2},
  pages={2716--2723},
  year={2022}
}

@inproceedings{ref76,
  title={Action-based contrastive learning for trajectory prediction},
  author={Halawa, Marah and Hellwich, Olaf and Bideau, Pia},
  booktitle={Proceedings of the European Conference on Computer Vision},
  pages={143--159},
  year={2022}
}

@inproceedings{ref77,
  title={Crossmodal transformer based generative framework for pedestrian trajectory prediction},
  author={Su, Zhaoxin and Huang, Gang and Zhang, Sanyuan and Hua, Wei},
  booktitle={Proceedings of the IEEE International Conference on Robotics and Automation },
  pages={2337--2343},
  year={2022},
  organization={IEEE}
}

@article{ref78,
  title={Bitrap: Bi-directional pedestrian trajectory prediction with multi-modal goal estimation},
  author={Yao, Yu and Atkins, Ella and Johnson-Roberson, Matthew and Vasudevan, Ram and Du, Xiaoxiao},
  journal={IEEE Robotics and Automation Letters},
  volume={6},
  number={2},
  pages={1463--1470},
  year={2021}
}

@ARTICLE{ref79,
  author={Zhang, Zhong and Zhou, Jianglin and Liu, Shuang and Xiao, Baihua},
  journal={IEEE Transactions on Multimedia}, 
  title={Completed Interaction Networks for Pedestrian Trajectory Prediction}, 
  year={2025},
  volume={27},
  pages={5119-5129}
}

@ARTICLE{ref80,
  author={Zhang, Zhengming and Ding, Zhengming and Tian, Renran},
  journal={IEEE Transactions on Image Processing}, 
  title={Decouple Ego-View Motions for Predicting Pedestrian Trajectory and Intention}, 
  year={2024},
  volume={33},
  pages={4716-4727}
}

@inproceedings{ref83,
  title={Distilling knowledge for short-to-long term trajectory prediction},
  author={Das, Sourav and Camporese, Guglielmo and Cheng, Shaokang and Ballan, Lamberto},
  booktitle={Proceedings of the IEEE/RSJ International Conference on Intelligent Robots and Systems (IROS)},
  pages={13001--13008},
  year={2024}
}

@inproceedings{ref85,
  title={Trajectory mamba: Efficient attention-mamba forecasting model based on selective ssm},
  author={Huang, Yizhou and Cheng, Yihua and Wang, Kezhi},
  booktitle={Proceedings of the IEEE/CVF Conference on Computer Vision and Pattern Recognition},
  pages={12058--12067},
  year={2025}
}

@inproceedings{ref86,
  title={Continuous locomotive crowd behavior generation},
  author={Bae, Inhwan and Lee, Junoh and Jeon, Hae-Gon},
  booktitle={Proceedings of the IEEE/CVF Conference on Computer Vision and Pattern Recognition},
  pages={22416--22431},
  year={2025}
}

@inproceedings{ref92,
  title={Feature pyramid networks for object detection},
  author={Lin, Tsung-Yi and Doll{\'a}r, Piotr and Girshick, Ross and He, Kaiming and Hariharan, Bharath and Belongie, Serge},
  booktitle={Proceedings of the IEEE/CVF Conference on Computer Vision and Pattern Recognition},
  pages={2117--2125},
  year={2017}
}

@inproceedings{ref93,
  title={Mvitv2: Improved multiscale vision transformers for classification and detection},
  author={Li, Yanghao and Wu, Chao-Yuan and Fan, Haoqi and Mangalam, Karttikeya and Xiong, Bo and Malik, Jitendra and Feichtenhofer, Christoph},
  booktitle={Proceedings of the IEEE/CVF conference on computer vision and pattern recognition},
  pages={4804--4814},
  year={2022}
}

@inproceedings{ref94,
  title={Dynamic multiscale graph neural networks for 3d skeleton based human motion prediction},
  author={Li, Maosen and Chen, Siheng and Zhao, Yangheng and Zhang, Ya and Wang, Yanfeng and Tian, Qi},
  booktitle={Proceedings of the IEEE/CVF conference on computer vision and pattern recognition},
  pages={214--223},
  year={2020}
}

@inproceedings{ref95,
  title={Visual point cloud forecasting enables scalable autonomous driving},
  author={Yang, Zetong and Chen, Li and Sun, Yanan and Li, Hongyang},
  booktitle={Proceedings of the IEEE/CVF Conference on Computer Vision and Pattern Recognition},
  pages={14673--14684},
  year={2024}
}

@inproceedings{ref96,
  title={Temporal pyramid network for pedestrian trajectory prediction with multi-supervision},
  author={Liang, Rongqin and Li, Yuanman and Li, Xia and Tang, Yi and Zhou, Jiantao and Zou, Wenbin},
  booktitle={Proceedings of the AAAI conference on artificial intelligence},
  volume={35},
  number={3},
  pages={2029--2037},
  year={2021}
}

@article{ref97,
  title={Multi-Scale Learnable Gabor Transform for Pedestrian Trajectory Prediction From Different Perspectives},
  author={Feng, Ang and Han, Cheng and Gong, Jun and Yi, Yang and Qiu, Ruiqi and Cheng, Yang},
  journal={IEEE Transactions on Intelligent Transportation Systems},
  year={2024},
  publisher={IEEE}
}

@inproceedings{ref98,
  title={Groupnet: Multiscale hypergraph neural networks for trajectory prediction with relational reasoning},
  author={Xu, Chenxin and Li, Maosen and Ni, Zhenyang and Zhang, Ya and Chen, Siheng},
  booktitle={Proceedings of the IEEE/CVF Conference on Computer Vision and Pattern Recognition},
  pages={6498--6507},
  year={2022}
}

@inproceedings{ref99,
  title={Mart: Multiscale relational transformer networks for multi-agent trajectory prediction},
  author={Lee, Seongju and Lee, Junseok and Yu, Yeonguk and Kim, Taeri and Lee, Kyoobin},
  booktitle={Proceedings of the European Conference on Computer Vision},
  pages={89--107},
  year={2024}
}

@article{ref100,
  title={Another vertical view: A hierarchical network for heterogeneous trajectory prediction via spectrums},
  author={Xia, Beihao and Wong, Conghao and Xu, Duanquan and Peng, Qinmu and You, Xinge},
  journal={IEEE Transactions on Pattern Analysis and Machine Intelligence},
  year={2025}
}

@inproceedings{ref101,
  title={Eigentrajectory: Low-rank descriptors for multi-modal trajectory forecasting},
  author={Bae, Inhwan and Oh, Jean and Jeon, Hae-Gon},
  booktitle={Proceedings of the IEEE/CVF International Conference on Computer Vision},
  pages={10017--10029},
  year={2023}
}

@inproceedings{ref102,
  author    = {Fu, Yuxiang and Yan, Qi and Wang, Lele and Li, Ke and Liao, Renjie},
  title     = {MoFlow: One-Step Flow Matching for Human Trajectory Forecasting via Implicit Maximum Likelihood Estimation based Distillation},
  booktitle   = {Proceedings of the IEEE/CVF Conference on Computer Vision and Pattern Recognition},
  year      = {2025}
}

@ARTICLE{ref104,
  author={Bae, Inhwan and Lee, Junoh and Jeon, Hae-Gon},
  journal={IEEE Transactions on Pattern Analysis and Machine Intelligence}, 
  title={Social Reasoning-Aware Trajectory Prediction via Multimodal Language Model}, 
  year={2025},
  pages={1-18}
}

@inproceedings{ref105,
  title={A unified environmental network for pedestrian trajectory prediction},
  author={Su, Yuchao and Li, Yuanman and Wang, Wei and Zhou, Jiantao and Li, Xia},
  booktitle={Proceedings of the AAAI Conference on Artificial Intelligence},
  volume={38},
  number={5},
  pages={4970--4978},
  year={2024}
}

@ARTICLE{ref106,
  author={Yang, Biao and Fan, Fucheng and Ni, Rongrong and Wang, Hai and Jafaripournimchahi, Ammar and Hu, Hongyu},
  journal={IEEE Transactions on Intelligent Transportation Systems}, 
  title={A Multi-Task Learning Network With a Collision-Aware Graph Transformer for Traffic-Agents Trajectory Prediction}, 
  year={2024},
  volume={25},
  number={7},
  pages={6677-6690}
}

@ARTICLE{ref107,
  author={Chen, Kai and Song, Xiao and Ren, Xiaoxiang},
  journal={IEEE Transactions on Circuits and Systems for Video Technology}, 
  title={Pedestrian Trajectory Prediction in Heterogeneous Traffic Using Pose Keypoints-Based Convolutional Encoder-Decoder Network}, 
  year={2021},
  volume={31},
  number={5},
  pages={1764-1775}
}

@ARTICLE{ref108,
  author={Yang, Biao and Wei, Zhiwen and Hu, Chuan and Cai, Yingfeng and Wang, Hai and Hu, Hongyu},
  journal={IEEE Transactions on Intelligent Transportation Systems}, 
  title={Real-Time Pedestrian Crossing Anticipation Based on an Action–Interaction Dual-Branch Network}, 
  year={2024},
  volume={25},
  number={12},
  pages={21021-21034}
}

@article{ref109,
  title={Non-probability sampling network based on anomaly pedestrian trajectory discrimination for pedestrian trajectory prediction},
  author={Liu, Quankai and Sang, Haifeng and Wang, Jinyu and Chen, Wangxing and Liu, Yulong},
  journal={Image and Vision Computing},
  volume={143},
  pages={104954},
  year={2024}

}

@ARTICLE{ref110,
  author={Xie, Bochen and Deng, Yongjian and Shao, Zhanpeng and Li, Youfu},
  journal={IEEE Transactions on Multimedia}, 
  title={EISNet: A Multi-Modal Fusion Network for Semantic Segmentation With Events and Images}, 
  year={2024},
  volume={26},
  pages={8639-8650}
}

@ARTICLE{ref111,
  author={Zhang, Jiaming and Liu, Huayao and Yang, Kailun and Hu, Xinxin and Liu, Ruiping and Stiefelhagen, Rainer},
  journal={IEEE Transactions on Intelligent Transportation Systems}, 
  title={CMX: Cross-Modal Fusion for RGB-X Semantic Segmentation With Transformers}, 
  year={2023},
  volume={24},
  number={12},
  pages={14679-14694}
}

@INPROCEEDINGS{ref112,
  author={Li, Yaowei and Quan, Ruijie and Zhu, Linchao and Yang, Yi},
  booktitle={Proceedings of the IEEE/CVF Conference on Computer Vision and Pattern Recognition}, 
  title={Efficient Multimodal Fusion via Interactive Prompting}, 
  year={2023},
  pages={2604-2613}
}

@ARTICLE{ref113,
  author={Liu, Yu and Liu, Zhijie and Yang, Zedong and Li, You-Fu and Kong, He},
  journal={IEEE Transactions on Intelligent Transportation Systems}, 
  title={Occlusion-Aware Diffusion Model for Pedestrian Intention Prediction}, 
  year={2026},
  volume={27},
  number={3},
  pages={3579-3593}
}

@article{ref114, 
title={Intention-Aware Diffusion Model for Pedestrian Trajectory Prediction}, 
volume={40},  
number={22}, 
journal={Proceedings of the AAAI Conference on Artificial Intelligence}, 
author={Liu, Yu and Liu, Zhijie and Ren, Xiao and Li, Youfu and Kong, He}, 
year={2026}, 
pages={18469-18477} }

@INPROCEEDINGS{ref115,
  author={Liu, Yu and Liu, Zhijie and Ren, Xiao and Li, You-Fu and Kong, He},
  booktitle={2025 IEEE 28th International Conference on Intelligent Transportation Systems }, 
  title={Intention Enhanced Diffusion Model for Multimodal Pedestrian Trajectory Prediction}, 
  year={2025},
  pages={2063-2068}
  }

@article{ref116,
  title={Pedestrian intention prediction for autonomous vehicles: A comprehensive survey},
  author={Sharma, Neha and Dhiman, Chhavi and Indu, Seema},
  journal={Neurocomputing},
  volume={508},
  pages={120--152},
  year={2022}
}

@article{ref117,
  title={Predicting pedestrian intentions with multimodal IntentFormer: A Co-learning approach},
  author={Sharma, Neha and Dhiman, Chhavi and Indu, Sreedevi},
  journal={Pattern Recognition},
  volume={161},
  pages={111205},
  year={2025}
}

@inproceedings{ref118,
  title={Tamformer: Multi-modal transformer with learned attention mask for early intent prediction},
  author={Osman, Nada and Camporese, Guglielmo and Ballan, Lamberto},
  booktitle={ICASSP 2023-2023 IEEE International Conference on Acoustics, Speech and Signal Processing (ICASSP)},
  pages={1--5},
  year={2023}
}

@article{ref119,
  title={Visual--motion--interaction-guided pedestrian intention prediction framework},
  author={Sharma, Neha and Dhiman, Chhavi and Indu, S},
  journal={IEEE Sensors Journal},
  volume={23},
  number={22},
  pages={27540--27548},
  year={2023}
}

@ARTICLE{ref120,
  author={Sharma, Neha and Dhiman, Chhavi and Indu, S.},
  journal={IEEE Transactions on Intelligent Vehicles}, 
  title={Progressive Contextual Trajectory Prediction With Adaptive Gating and Fuzzy Logic Integration}, 
  year={2024},
  volume={9},
  number={11},
  pages={6960-6970}
}

@article{ref122,
  title={Cross-Modal Pedestrian Behavior Prediction: A Dual-Task Approach With Progressive Denoising Attention and CVAE},
  author={Sharma, Neha and Dhiman, Chhavi and Indu, Sreedevi},
  journal={IEEE Transactions on Intelligent Transportation Systems},
  year={2025}
}

@inproceedings{ref123,
  title={Multimodal Transformer Networks for Pedestrian Trajectory Prediction.},
  author={Yin, Ziyi and Liu, Ruijin and Xiong, Zhiliang and Yuan, Zejian},
  booktitle={IJCAI},
  pages={1259--1265},
  year={2021}
}

@INPROCEEDINGS{ref124,
  author={Yin, Zeya and Lai, Tin and Barcelos, Lucas and Jacob, Jayadeep and Li, Yonghui and Ramos, Fabio},
  booktitle={Proceedings of the IEEE International Conference on Robotics and Automation}, 
  title={Diverse Motion Planning with Stein Diffusion Trajectory Inference}, 
  year={2025},
  pages={15610-15616}
}

\begin{IEEEbiography}[{\includegraphics[width=1in,height=1.24in,clip,keepaspectratio]{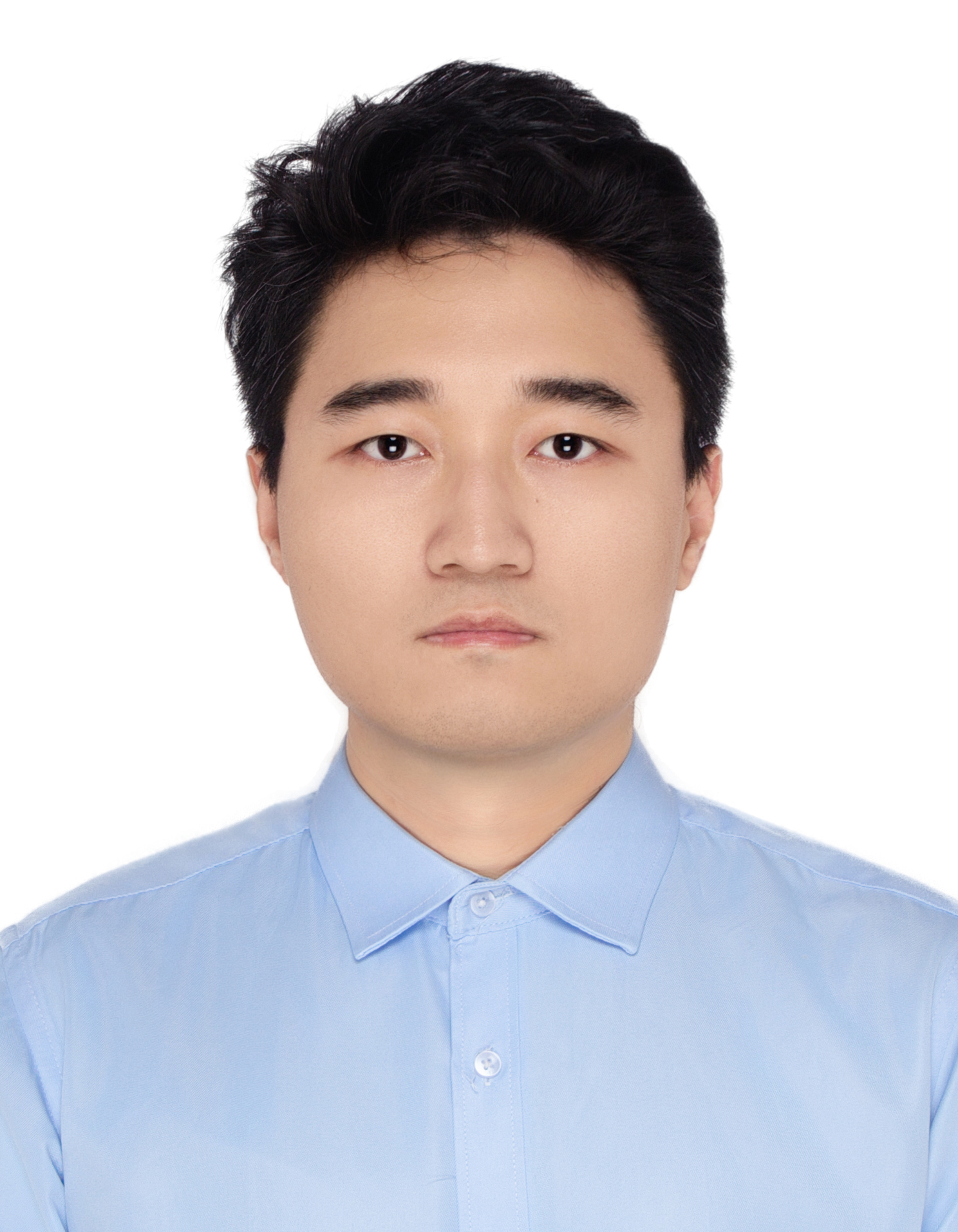}}]{Yu Liu}
received the Bachelor's degree in Process Equipment and Control Engineering from Chongqing University of Technology, Chongqing, China, and the Master's degree in Aerospace Engineering from Nagoya University, Nagoya, Japan. He is currently pursuing the Ph.D. degree with the Department of Mechanical Engineering, City University of Hong Kong (CityU), Hong Kong SAR, China, and also with the Shenzhen Key Laboratory of Control Theory and Intelligent Systems, Southern University of Science and Technology (SUSTech), Shenzhen, China. His current research interests include deep learning, motion prediction, and intelligent vehicles.
\end{IEEEbiography}	

\begin{IEEEbiography}[{\includegraphics[width=1in,height=1.25in,clip,keepaspectratio]{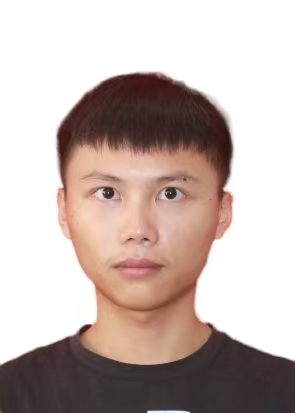}}]{Ming Huang }
received the B.S. degree in Electronic Information Engineering from the University of South China, Hengyang, China, in 2025. He is currently pursuing the M.S. degree at the School of Automation and Intelligent Manufacturing, Southern University of Science and Technology (SUSTech), Shenzhen, China. His research interests include sound source localization algorithms and the design of robotic auditory systems.
\end{IEEEbiography}

\begin{IEEEbiography}[{\includegraphics[width=1in,height=1.25in,clip,keepaspectratio]{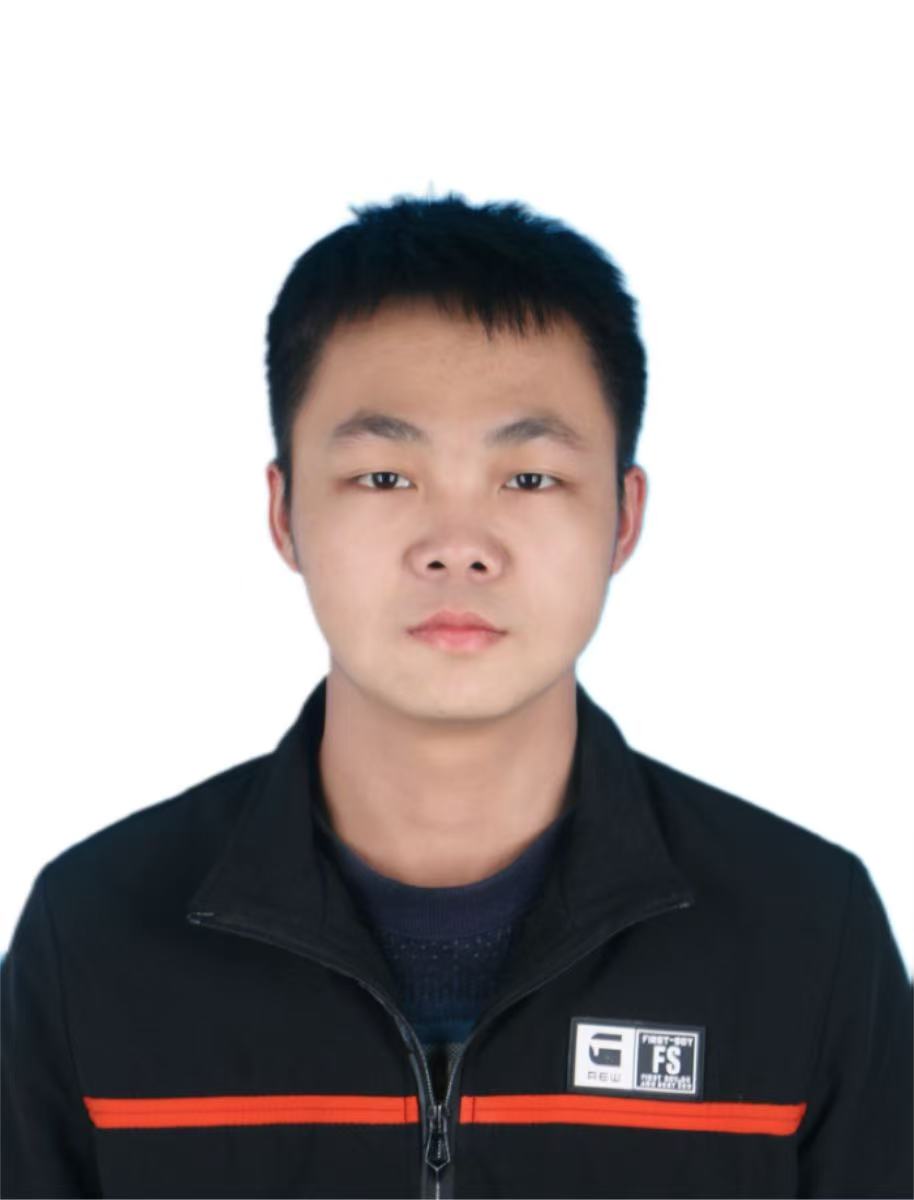}}]{Xiao Ren}
received the Bachelor's degree in Electronic Information Engineering from Hubei University of Automotive Industry, Shiyan, China, and a Master's degree in Electronic Information Engineering from Beijing University of Information Science and Technology (BUIST), Beijing, China. He is currently working towards a Ph.D. degree at Southern University of Science and Technology (SUSTech) in Shenzhen, China. His current research interests include robot perception and localization, and 3D reconstruction. 
\end{IEEEbiography}

\begin{IEEEbiography}[{\includegraphics[width=1in,height=1.25in,clip,keepaspectratio]{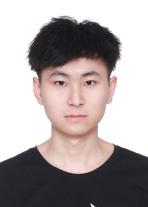}}]{Zhijie Liu}
    received the B.S. degree in Measurement and Control Technology and Instrument from Harbin Engineering University, Harbin, China in 2023. He is currently pursuing the M.S. degree at the School of System Design and Intelligent Manufacturing, Southern University of Science and Technology (SUSTech), Shenzhen, China.
    His research interests include nonlinear control, underactuated systems, optimal control, observer design, and high-order fully actuated system approaches.
\end{IEEEbiography}

\begin{IEEEbiography}[{\includegraphics[width=1in,height=1.25in,clip,keepaspectratio]{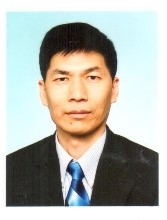}}]{Youfu Li}
  received the PhD degree in robotics from the Department of Engineering Science, University of Oxford in 1993. From 1993 to 1995 he was a research staff in the Department of Computer Science at the University of Wales, Aberystwyth, UK. He joined City University of Hong Kong in 1995 and is currently professor in the Department of Mechanical Engineering. His research interests include robot sensing, robot vision, and visual tracking. In these areas, he has published over 400 papers including over 180 SCI listed journal papers. Dr Li has received many awards in robot sensing and vision including IEEE Sensors Journal Best Paper Award by IEEE Sensors Council, Second Prize of Natural Science Research Award by the Ministry of Education, 1st Prize of Natural Science Research Award of Hubei Province, 1st Prize of Natural Science Research Award of Zhejiang Province, China. He was on Top 2\% of the world’s most highly cited scientists by Stanford University, 2020, 2021 and Career Long. He has served as an Associate Editor for IEEE Transactions on Automation Science and Engineering (T-ASE), Associate Editor and Guest Editor for IEEE Robotics and Automation Magazine (RAM), and Editor for CEB, IEEE International Conference on Robotics and Automation (ICRA). He is a Fellow of the IEEE.
\end{IEEEbiography}	

\begin{IEEEbiography}[{\includegraphics[width=1in,height=1.25in,clip,keepaspectratio]{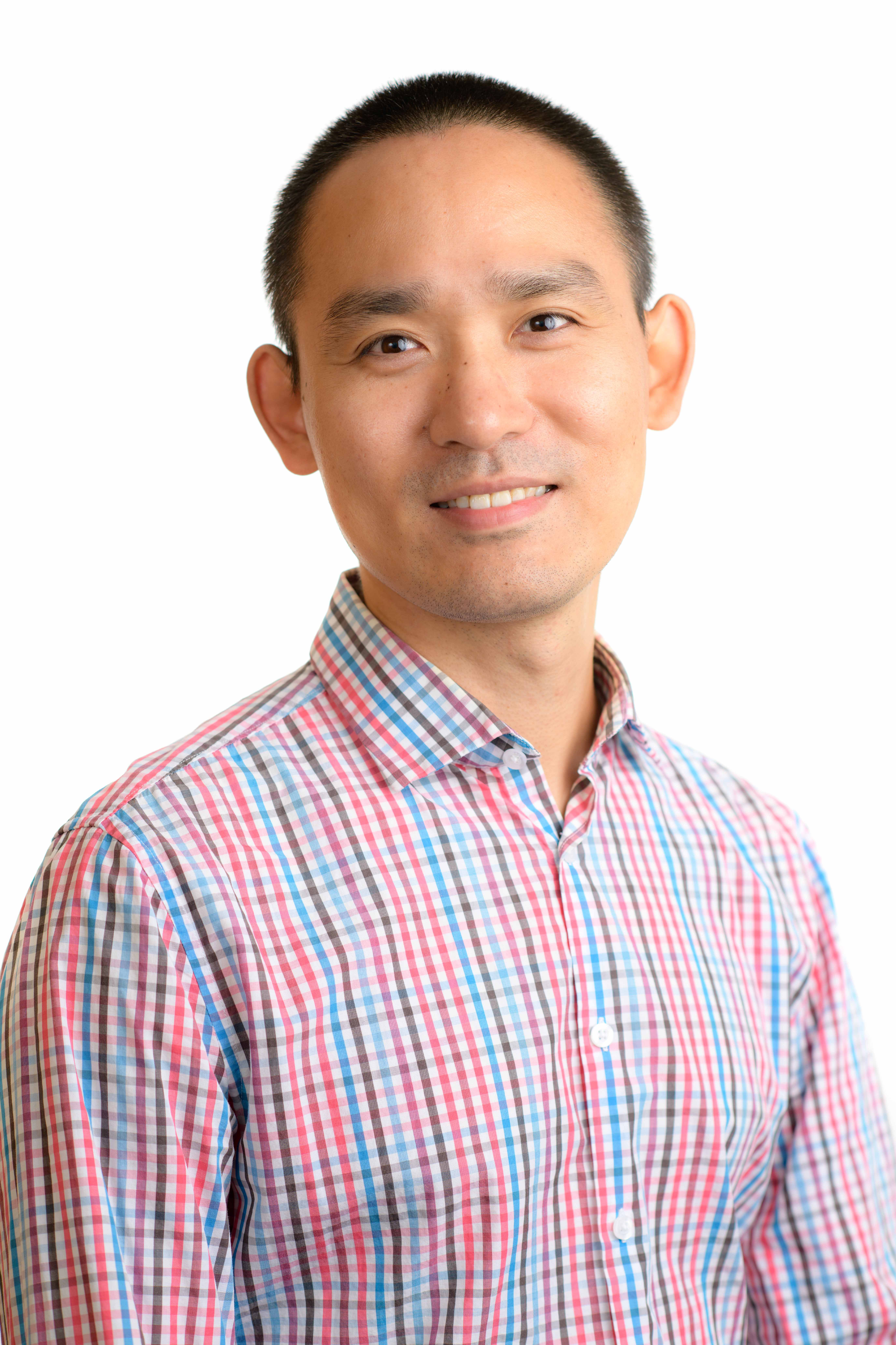}}]{He Kong}
  received the Ph.D. degree in Electrical Engineering from the University of Newcastle, Australia, respectively. He was a Research Fellow at the Australian Centre for Field Robotics, the University of Sydney, Australia, during 2016–2021. In early 2022, he joined the Southern University of Science and Technology, Shenzhen, China, where he is currently an Associate Professor. His research interests include active multi-modal perception, robot audition, state estimation, and control applications. He is currently serving on the editorial board of \textit{IEEE Robotics and Automation Letters}, \textit{IEEE Robotics and Automation Magazine}, \textit{International Journal of Adaptive Control and Signal Processing}, \textit{the IEEE Sensor Letters}, \textit{The Proc. of the Institution of Mechanical Engineers, Part I: Journal of Systems and Control Engineering}, etc. He has served as an Associate Editor on the IEEE Control System Society Conference Editorial Board as well as for flagship conferences of the IEEE Robotics and Automation Society.
\end{IEEEbiography}

\end{document}